# A Decision-Focused Neural Optimization Framework for Personalized Route Reproduction from Vehicle Trajectories

**Gyeongjun Kim**
Graduate student
Department of Smart City, Chung-Ang University, Seoul, Korea
84 Heukseok-ro, Dongjak-gu, Seoul 156-756, Korea
kkjunn7033@cau.ac.kr

**Yeseul Kang**
Graduate student
Department of Urban Engineering, Chung-Ang University, Seoul, Korea
84 Heukseok-ro, Dongjak-gu, Seoul 156-756, Korea
dptmf0307@cau.ac.kr

**Keemin Sohn*** Corresponding author
Professor
Department of Urban Engineering, Department of Smart City, Chung-Ang University, Seoul, Korea
84 Heukseok-ro, Dongjak-gu, Seoul 156-756, Korea
kmsohn@cau.ac.kr

## Abstract

To reproduce observed routes at the individual level, this study revisits the fundamental premise that travelers aim to minimize their perceived costs. Once these subjective link travel costs are identified, the route choice problem reduces to a shortest-path search without the need to enumerate alternative routes. We propose a neural pipeline that includes a perception model that embeds context covariates, which comprises individual characteristics, trip-specific attributes, and network-level traffic states, into the perceived link costs. A constrained optimization (CO) layer, which determines the shortest path (SP) based on these estimated costs, follows the perception encoder. To enable end-to-end training, we employ decision-focused learning to align the predicted shortest paths with observed routes. The implicit maximum likelihood estimation (iMLE) provides an approximate gradient of the loss function that contains the non-differentiable CO layer. Furthermore, a regularization term added to the loss function ensures proper scale of perceived link travel costs. Empirical evaluations demonstrate that the proposed framework outperforms baseline route choice models in path reproduction. The framework's capability of retrieving perceived travel costs offers plausible interpretation for route choice motives at the individual level, providing insights into why drivers exhibit heterogeneous behavior under identical traffic conditions.

# 1. Introduction

The route choice behavior has long been studied in transportation research fields (Ben-Akiva and Lerman, 1985; Bovy and Stern, 1990). Empirical evidence consistently demonstrates that travelers exhibit significant heterogeneity, choosing diverse routes even under identical traffic conditions (Jan et al., 2000; Papinski et al., 2009; Levinson and Zhu, 2013). Empirical studies further reveal that the proportion of travelers opting for the objectively minimum travel time path, even when accounting for network congestion, typically ranges between only 40% and 60% (Ramírez et al., 2021; Zhu and Levinson, 2015). To account for this behavioral diversity, discrete choice models (DCMs) have been extensively employed.

However, traditional DCMs encounter significant computational hurdles, particularly regarding the generation of choice sets. As enumerating all feasible paths in a large-scale network is practically infeasible, model estimation often relies on a pre-defined subset of routes, which entails potential biases (Ben-Akiva and Bierlaire, 1999; Bekhor et al., 2001; Frejinger et al., 2009). While the recursive logit model addresses the path enumeration problem through a link-based formulation (Fosgerau et al., 2013), it remains constrained by the independence of irrelevant alternatives (IIA) property. Consequently, logit-based route choice models struggle to accurately reproduce rational route choice behavior when candidate paths exhibit high degrees of spatial overlap. At the network level, Fosgerau et al. (2022) proposed the perturbed utility route choice (PURC) model that is also free from both the alternative enumeration and the IIA challenge. The model calibration method was also extended to utilize individual trajectories (Fosgerau et al., 2024). The extension, however, is not to retrieve the individual-level perceived link-cost from observed trajectories but to calibrate the model to retrieve link traffic volumes.

In this study, we revisit the perceived-cost minimization principle: conditional on a traveler-specific perceived link costs, the realized route is represented as the minimum cost path. Although this principle is a fundamental behavioral premise in route choice and traffic assignment, existing models typically use it indirectly through likelihood-based route probabilities (Frejinger et al., 2009), recursive value functions (Fosgerau et al., 2013), perturbed utility flow allocation (Fosgerau et al., 2022), or latent choice-set perception (Yao and Bekhor, 2022). To the best of our knowledge, few studies have directly considered an individual-level perceived link costs by embedding the full-network shortest-path (SP) finding operator in the learning objective and matching its resulting path to each observed trajectory. Including the SP finding operation makes it difficult to calibrate the model via maximizing the likelihood. Kim et al. (2026) sheds light on utilizing the principle directly for the model calibration, wherein a probit choice model was calibrated directly by the random utility maximization (RUM) theory at the individual level, not depending on the likelihood maximization that demands the evaluation of probit choice function. Whereas this previous study approximates the 'argmax' operation with a differentiable function, the present study keeps the shortest-path layer discrete in the forward pass and uses an iMLE-based surrogate gradient for backpropagation.

We devise a modeling framework to retrieve the latent perceived costs that lead to the observed paths. The proposed framework encodes context covariates, including individual characteristics, trip-specific attributes, and network-wide traffic states, into perceived link costs. A constrained optimization (CO) layer to determine the minimum-cost path based on the principle above, effectively reproducing the observed trajectory, is then attached to the encoder. This modeling framework is free from preparing an alternative route set that is prerequisite for other path-based DCMs. It should be noted that the proposed framework is not a likelihood-based route choice probability estimator. Rather, the framework is an intelligent route-reproduction framework that retrieves individual-specific perceived link costs and evaluates them through a full-network SP layer.

While the proposed end-to-end pipeline is conceptually intuitive and computationally efficient during inference, its training phase presents non-trivial mathematical challenges. Specifically, backpropagating gradients through a discrete SP solver requires addressing the non-differentiability. Unlike conventional supervised learning, which directly maps inputs to labels, decision-focused learning seeks to align the solution of an optimization problem with the revealed output. To overcome the gradients' vanishing or exploding nature in such neural-optimization pipelines, recent advancements have introduced techniques to enable backpropagation through CO layers (Wilder et al., 2019; Elmachtoub and Grigas, 2022; Mandi et al., 2024). In this study, we propose a robust calibration methodology by employing the implicit maximum likelihood estimation (iMLE) method. The iMLE method provides an approximate gradient of the loss function that contains non-differentiable CO layers (Vlastelica et al., 2020; Niepert et al., 2021), thereby allowing for the calibration of the end-to-end route reproduction model and the retrieval of perceived travel costs, while maintaining the structural integrity of the underlying optimization problem.

Another critical challenge in estimating perceived link travel costs is ensuring the unique set of perceived travel times. Since multiple sets of perceived link costs can yield the identical SP solution, the estimation process is inherently ill-posed. To alleviate this non-uniqueness problem, we introduce a regularization term to the loss function when training, which ensures cost-scale anchoring. This regularization effectively anchors the estimated costs at the average travel times over trips, resulting in perceived link travel costs that align with the scale of observed travel costs. By penalizing extreme deviations, the proposed approach provides a robust and interpretable set of latent costs that maintains physical coherence.

The contributions of this study are three-fold. **First**, an individual-level personalized route reproduction framework is proposed. We devise a framework that operates directly on perceived link travel costs, eliminating the need for explicitly enumerating candidate routes. **Second**, decision-focused learning (Mandi et al., 2024) is adapted to calibrate the proposed route-reproduction framework. The iMLE-based surrogate gradient enables end-to-end training (Niepert et al., 2021), while preserving the discrete SP solver in the forward pass. **Third**, the retrieval of perceived travel costs can provide a mechanistic explanation for personalized route choice behavior. Unlike existing discrete choice models that rely on aggregate stochasticity, our framework offers granular interpretability by identifying the latent cost structures that drive heterogeneous decisions under identical traffic conditions.

The remainder of this paper is organized as follows. Section 2 provides a comprehensive literature review, focusing on the evolution of route choice models within the discrete choice modeling (DCM) framework and identifying the gaps our research aims to bridge. This section also introduces several similar studies in the field of transportation and highlights the superiority of the present study. Section 3 formalizes the proposed modeling framework, details the calibration process via decision-focused learning, and introduces the regularization scheme designed to address the cost-scale ambiguity. Section 4 describes the datasets utilized for the analysis and outlines the experimental configuration. Section 5 presents empirical results and discussions, where the proposed framework is validated against real-world vehicle trajectories to assess its performance and interpretability. Finally, Section 6 concludes the study by summarizing the key findings and discussing potential avenues for future research.

## 2. Related work

### 2.1 Discrete choice models of route choice

Route choice has traditionally been modeled within the discrete choice framework, which broadly distinguishes between path-based and link-based formulations. Path-based models represent route choice as a selection among complete origin–destination paths and are attractive because their behavioral interpretation is straightforward at the route level (Ben-Akiva and Bierlaire, 1999; Prato, 2009). However, path-based models face two persistent challenges. They require a finite candidate path set to be generated in advance, and their estimates can be sensitive to how that choice set is constructed. A second difficulty is that candidate routes often overlap substantially in real networks, which violates the independence assumptions of simple multinomial logit specifications and can lead to unrealistic substitution patterns across. To mitigate this issue, overlap-correction models such as the path size logit have been widely adopted, while later studies also explored richer choice-set generation and path representation strategies (Ben-Akiva and Bierlaire, 1999; Frejinger et al., 2009; Marra and Corman, 2020).

These limitations motivated the development of link-based model, which formulate route choice as a sequence of local link decisions on the network rather than as a single selection from a pre-enumerated path set. Recursive logit model (Fosgerau et al., 2013) is especially influential because it avoids explicit path enumeration, is equivalent to a multinomial logit model with an unrestricted path set, and provides a tractable framework for estimation and prediction on large networks. Subsequent extensions, including the nested recursive logit and discounted recursive logit, further improved the flexibility of the link-based framework and demonstrated its usefulness in diverse route choice settings and travel modes (Mai et al., 2015; Oyama and Hato, 2017; Zimmermann et al., 2017; Meyer de Freitas et al., 2019). Nevertheless, most existing discrete route choice models still rely on relatively simple linear utility specifications and link-additive representations, which limit their ability to capture nonlinear effects, richer trip context, and traveler-specific heterogeneity in perceived network conditions

### 2.2 Learning-based route choice

Beyond econometric and behavioral formulations, recent studies have increasingly explored learning-based approaches to relax the restrictive functional forms of classical route choice models and to better exploit rich trajectory and network data (Cantarella and de Luca, 2005; Wang et al., 2020; Zhao and Liang, 2023). On the path-based side, deep neural network variants of path size logit have been proposed to replace linear utility specifications with more flexible nonlinear mappings while retaining candidate-set-based route probabilities (Marra and Corman, 2021). These models improve representational flexibility, but they still inherit the structural dependence of path-based approaches on route generation and candidate-set coverage.

On the link-based side, inverse reinforcement learning and related sequential learning frameworks have been used to model route choice as a policy or reward-learning problem. Zhao and Liang (2023), for example, adopted an adversarial inverse reinforcement learning (AIRL) for route choice that learns a context-dependent reward function from trajectory data. Sequence-based route choice and path-inference models such as RoutesFormer further show that learned representations can improve predictive accuracy by exploiting longer-range dependency structures in observed trajectories (Qiu et al., 2024). These studies demonstrate the value of flexible, learned route-generation mechanisms and provide strong empirical baselines for comparison. However, most learning-based route choice methods remain focused on OD-conditional route distributions, link-choice policies, or latent reward functions. They do not attempt the behavioral interpretation as a driver's perceived link cost distribution.

### 2.3 Similar Studies with Latent Travel Costs

The perturbed utility route choice (PURC) model proposed by Fosgerau et al. (2022) is especially relevant because it also avoids explicit choice-set generation and operates directly on the full network. PURC represents route choice through a perturbed utility formulation and predicts network-consistent link-use outcomes by solving a convex optimization problem. Unlike recursive logit, it is not subject to the standard IIA limitation associated with logit-based path or link choice models. More recent trajectory-based extensions further allow PURC-type models to be estimated from individual observed paths (Fosgerau et al., 2024). More recently, Rosenblad et al. (2026) further develops PURC estimation by proposing a convex inverse-estimation framework using observed link-use proportions constructed from vehicle trajectory data.

These PURC studies provide the closest full-network statistical benchmark for the present study. Nevertheless, their prediction output is an expected link-use or flow vector at each origin destination (OD) pair, rather than a deterministic reproduced path for each individual trip. Moreover, PURC is not designed to retrieve individual-specific perceived link costs that rationalizes each driver's realized trajectory under a SP rule. In this sense, PURC and its inverse-estimation extensions are highly relevant but differ from the target output and interpretation of the proposed framework.

Recently, a parallel line of studies has integrated an optimization layer into a neural pipeline (Liu et al., 2023), into an objective function for model calibration (Sun and Arslan, 2025), or into a full-network route reproduction formulation without explicit path enumeration (Fosgerau et al., 2022). These studies share the view that observed route or network outcomes can be better reproduced when the model uses latent or perturbed link costs rather than raw observed travel times alone. A key idea is that an optimization solver can act as a behavioral or operational layer that maps latent costs into observable route, tour, or flow outcomes.

Liu et al. (2023) learn latent cost structures under traffic assignment constraints by adding a user-equilibrium (UE) traffic assignment layer to a neural pipeline to fit multi-day link flow observations. This shows that unobserved cost components can be inferred from demand and context data. However, the resulting travel cost is essentially a population-level cost surface shared across travelers, and thus it does not directly explain why a specific individual chooses a particular realized route. They transform the UE traffic assignment problem into a variational inequality (VI) formulation and suggest a fixed-point algorithm for solving the problem. A gradient-based algorithm is adopted for training, and the gradient is derived via the implicit differentiation theorem. Deriving the gradient, however, requires handling a large-dimensional Jacobian, which entails a substantial computational burden.

Sun and Arslan (2025) suggest an experienced learning approach for solving a traveling salesman problem (TSP). They learn data-driven transition weights that adjust edge costs so that a routing solver generates solutions consistent with observed high-quality tours in last-mile delivery. While the learned adjustments improve realism, they are designed as a shared set of transition weights that can be reused across operations, rather than as personalized perceived link costs. In short, latent cost learning and perturbed utility modeling have progressed toward calibrating hidden network cost components from observed data, but existing approaches typically target aggregate link flows, expected link-use probabilities, and shared routing weights. This limits their ability to interpret why a particular individual selected a particular realized trajectory.

In contrast, the present study targets individual-level route reproduction by learning trip-specific perceived link costs from exogenous traveler, trip, and network-context variables and then computing the corresponding minimum perceived-cost path. The proposed framework directly uses the SP solution as the supervised object in calibration: the encoder is trained so that the path induced by the learned perceived costs matches the observed trajectory. In addition, the iMLE method employed in this study provides a surrogate gradient by repeatedly evaluating the discrete SP layer, thereby avoiding explicit differentiation of a large-dimensional equilibrium or assignment system.

### 2.4 Decision-Focused Learning and Differentiable Constrained Optimization

Many practical decision problems are solved under uncertainty because the optimization model depends on quantities that are not directly observed and therefore must be estimated from data, such as costs, demands, or rewards (Mandi et al., 2024). A common strategy is the predict-then-optimize framework, in which a predictive model first estimates these quantities and an optimization model then computes the final decision (Bertsimas and Kallus, 2020; Mandi et al., 2024). However, better prediction accuracy does not necessarily lead to better decisions, because some estimation errors have little effect on the optimal solution while others can change the selected solution substantially even when the numerical error is small (Wilder et al., 2019; Elmachtoub and Grigas, 2022).

Decision-focused learning was introduced to address this mismatch by training predictive models according to the quality of the decisions they induce, rather than according to intermediate prediction error alone (Wilder et al., 2019; Elmachtoub and Grigas, 2022). Under this perspective, prediction and optimization are combined into a single end-to-end framework so that model parameters are updated based on downstream task performance (Wilder et al., 2019; Mandi et al., 2024). This shift is especially useful in constrained and combinatorial decision problems, where the practical objective is to obtain a high-quality decision and not merely an accurate estimate of latent model inputs (Demirović et al., 2019; Elmachtoub and Grigas, 2022; Mandi et al., 2024).

A key technical challenge, however, is that many downstream optimization problems are discrete and non-differentiable, which prevents ordinary backpropagation from being applied directly through the solver (Amos and Kolter, 2017; Vlastelica et al., 2020). Recent work on differentiable CO has addressed this issue by developing surrogate gradients, continuous relaxations, and perturbation-based methods that make it possible to train predictive models through optimization layers (Vlastelica et al., 2020; Niepert et al., 2021; Mandi et al., 2024). Among these approaches, perturbation-based methods such as iMLE are particularly appealing because they preserve the original discrete optimization problem in the forward pass while providing usable gradient estimates for end-to-end learning in the backward pass (Niepert et al., 2021). These developments provide an important methodological foundation for learning systems in which model outputs are ultimately evaluated through structured decisions rather than by prediction error alone (Mandi et al., 2024).

# 3. Methodology

## 3.1 Behavioral Assumptions

The proposed route reproduction framework is established upon two fundamental behavioral assumptions. The first concerns the derivation of subjective link-level perceptions, and the second addresses the path-selection mechanism based on those perceptions. Distinguishing these two processes clarifies how the proposed model deviates from conventional stochastic route choice formulations.

**Assumption 1 (Heterogeneous Link-Level Cost Perception):**

In the proposed framework, the perceived cost of link $a$ for traveler $n$ is modeled as $C_n^a = e^{y_n^a}$ and $y_n^a \sim N(\mu_n^a,\ \sigma_n^a)$, where $(\mu_n^a,\ \sigma_n^a)$ is modeled as a function of individual characteristics, trip-specific attributes, and network-level traffic states. It allows the latent cost distribution to be a lognormal and to vary across both travelers and links. This reparameterization scheme is the key departure of the proposed route choice model.

**Assumption 2 (Trip-level shortest-path finding on perceived costs):**

Conditional upon the perceived link costs, the traveler is assumed to follow a deterministic cost-minimization objective. The path perceived cost is the linear summation of its constituent link costs. Consequently, the observed route is treated as the solution to a shortest-path optimization problem.

Collectively, these assumptions underpin the architecture of the proposed model. By bypassing probabilistic choice modeling in favor of a deterministic solver acting on latent perceived costs, the model directly mimics the intrinsic route-choosing process of travelers rather than deriving the choice probability. The subsequent sections address the mathematical complexities involved in calibrating this framework of neural-optimization pipeline.

## 3.2 Problem set-ups

Let the directed road network be represented by a graph $G = (V, A)$, where $V$ and $A$ denote the sets of nodes and directed links, respectively. We use $a \in A$ to index links and $n \in \{1, \dots, N\}$ to index trips. Each trip $n$ is associated with an origin $o_n \in V$, a destination $d_n \in V$, a departure-time index $\boldsymbol{x}_n^{depT} \in \{1, \dots, T\}$, and origin- and destination-specific context variables ($\boldsymbol{x}_n^o\ \boldsymbol{x}_n^d$).

For each trip $n$ and link $a$, the model assigns a personalized perceived link cost $\hat{C}_{a,n}$. Each trip $n$ therefore has its own perceived link-cost vector ($\widehat{\boldsymbol{C}}_n \in \mathbb{R}^{|A|}$), with each link cost characterized by the trip maker's perceived travel cost ($\hat{C}_{a,n}$): $\widehat{\boldsymbol{C}}_n = [\hat{C}_{a,n}]_{a \in A}$.

Following the two behavioral assumptions in the previous subsection, we model route generation directly through the shortest path finding mechanism rather than computing choice probabilities.

Throughout the paper, $r_n$ denotes the observed route for trip $n$, whereas $\hat{r}_n$ denotes the route generated by the model. The learning objective is to align the generated route $\hat{r}_n$ with the observed route $r_n$, which leads to decision-focused learning while the perceived link costs remain latent.

Let $\mathbf{x}_n$ denote the covariate vector for trip $n$. Specifically, we write $\mathbf{x}_n = (\boldsymbol{x}_n^{person}, \boldsymbol{x}_n^{trip}, \boldsymbol{x}_n^{net})$, where $\boldsymbol{x}_n^{person}$ denotes traveler attributes, $\boldsymbol{x}_n^{trip}$ trip-context variables, and $\boldsymbol{x}_n^{net}$ network state information. The trip context variables ($\boldsymbol{x}_n^{trip}$) contain spatio-temporal trip properties such as departure time ($\boldsymbol{x}_n^{depT}$) and origin- and destination-specific context variables ($\boldsymbol{x}_n^{o}, \boldsymbol{x}_n^{d},$) The network state information ($\boldsymbol{x}_n^{net}$) includes static and dynamic traffic states ($\boldsymbol{x}_n^{static_net}, \boldsymbol{x}_n^{dynamic_net}$). The encoder model maps $\mathbf{x}_n$ to the parameters of a perceived travel cost distribution and can be calibrated by the reparameterization scheme (Kingma and Welling, 2014). Table 1 summarizes core notations in the present study.

Table 1. Notation descriptions

| Notation | Description |
|---|---|
| $\boldsymbol{G} = (\boldsymbol{V}, \boldsymbol{A})$ | Directed road network graph with node set $V$ and link set $A$ |
| $\boldsymbol{a}$ | Link index, $a \in A$ |
| $\boldsymbol{n}$ | Trip index, $n = 1, \dots, N$ |
| $\boldsymbol{x_n^{person}}$ | Individual-specific variables of trip $n$ |
| $\boldsymbol{x_n^o}, \boldsymbol{x_n^d}$ | Origin- and destination-specific variables of trip $n$ |
| $\boldsymbol{x_n^{depT}}$ | Departure time index of trip $n$ |
| $\boldsymbol{x_n^{static_net}}, \boldsymbol{x_n^{dynamic_net}}$ | Static and dynamic network states of trip $n$ |
| $\boldsymbol{x_n}$ | Exogenous input vector for trip $n$, collecting individual characteristics, trip-specific attributes, and network-level traffic states $[\boldsymbol{x_n^{person}}, \boldsymbol{x_n^{depT}}, \boldsymbol{x_n^o}, \boldsymbol{x_n^d}, \boldsymbol{x_n^{static_net}}, \boldsymbol{x_n^{dynamic_net}}]$ |
| $\boldsymbol{\widehat{C}_n}$ | Sampled perceived link-cost vector for trip $n$ |
| $\boldsymbol{\mu_n}, \boldsymbol{\Sigma_n}$ | Mean vector and covariance matrix of perceived link costs for trip $n$ |
| $\boldsymbol{f_\theta}(\cdot)$ | Neural encoder parameterized by $\boldsymbol{\theta}$ |
| $\boldsymbol{\varepsilon}$ | Perturbation noise used in perturb-and-MAP sampling |
| $\boldsymbol{r_n}$ | Observed route for trip $n$, represented as a link-incidence vector |
| $\boldsymbol{\hat{r}_n}$ | Predicted route for trip $n$ obtained from the shortest-path layer |

## 3.3 Modeling Framework

### 3.3.1 Establishing model architecture

Fig. 1 illustrates the end-to-end architecture that maps covariate inputs to perceived link costs and then solves a SP problem based on the costs to derive an output path. The final output is thus the shortest path, and the difference between this path and the ground truth is minimized for training the model. Although the architecture is conceptually simple, the main difficulty lies in training the model. The CO layer to find the shortest path is included in the model pipeline but is non-differentiable. Decision-focused learning can match the solution of a CO layer with the ground truth. The iMLE approach (Niepert et al., 2021) backpropagates the CO layer within the end-to-end model pipeline.

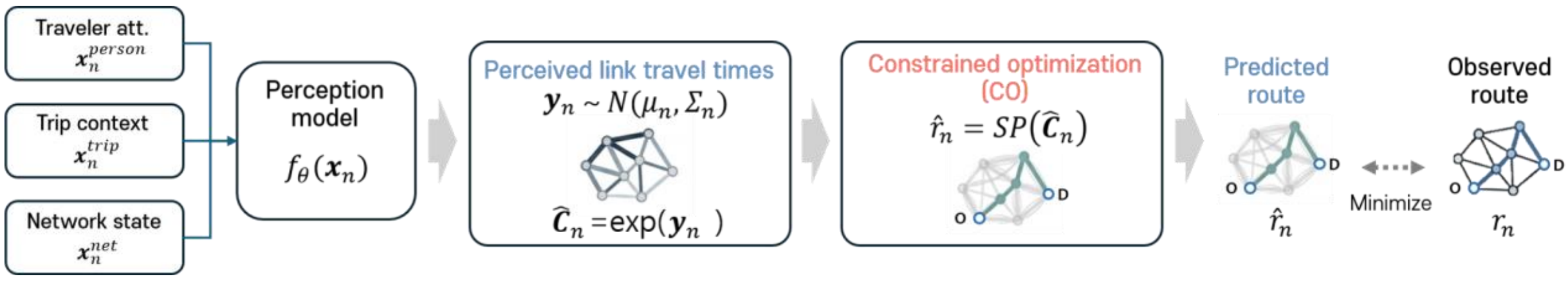

Fig. 1 Model architecture for the proposed modelling framework

The forward pass to evaluate the proposed model has a sampling procedure as denoted in Eq (1). The sampling is to get a perceived link cost vector ($\hat{\boldsymbol{C}}_n$). The perception model [$f_{\boldsymbol{\theta}}(\boldsymbol{x}_n)$] to encoded context covariates into perceived costs is approximated by a neural network and returns the mean ($\mu_n \in \mathbb{R}^{|A|}$) and variance ($\Sigma_n \in \mathbb{R}^{|A| \times |A|}$) of the Gaussian distribution. $\Sigma_n$ is assumed as a diagonal matrix. To make perceived travel costs positive, the exponential function is taken to samples from the encoder. This action reflects that perceived link travel times follow the lognormal distribution.

$$(\mu_n, \Sigma_n) = f_{\boldsymbol{\theta}}(\boldsymbol{x}_n),\ \ \boldsymbol{y}_n \sim \mathcal{N}_{\boldsymbol{\theta}}(\mu_n, \Sigma_n),\ \ \hat{\boldsymbol{C}}_n = e^{\boldsymbol{y}_n},\ \hat{r}_n = SP(\hat{\boldsymbol{C}}_n) \qquad (1)$$

Based on this reparameterization scheme (Kingma and Welling, 2014), a perceived link-cost vector $\hat{\boldsymbol{C}}_n$ is sampled and passed to the shortest-path (SP) layer. Conditional on $\hat{\boldsymbol{C}}_n$, the forward SP prediction is deterministic as follows

$$\hat{r}_n = SP\ (\hat{\boldsymbol{C}}_n) = \underset{r \in \mathcal{P}_n}{arg\,min}\, \hat{\boldsymbol{C}}_n^{\top} r \qquad (2)$$

### 3.3.2 Setting up the original loss function

The original decision loss measures the discrepancy between the observed route $r_n$ and the predicted route $\hat{r}_n$. In this study, the individual route-matching loss is defined as the complement of link-based intersection-over-union (IOU):

$$\ell(r_n, \hat{r}_n) = 1 - \frac{r_n^{\top} \hat{r}_n}{\mathbf{1}^{\top} r_n + \mathbf{1}^{\top} \hat{r}_n - r_n^{\top} \hat{r}_n} \qquad (3)$$

Thus, the original decision-focused objective can be written as Eq. (4).

$$L(\boldsymbol{\theta}) = \frac{1}{N} \sum_{n=1}^{N} L_n(\boldsymbol{\theta}), \text{ where } L_n(\boldsymbol{\theta}) = \mathbb{E}_{\hat{C}_n \sim \mathcal{N}(\mu_n, \Sigma_n)}[\ell(r_n, \hat{r}_n)] \qquad (4)$$

However, this objective cannot be directly optimized by ordinary backpropagation. The SP operator in Eq. (2) maps a continuous cost vector to a discrete path-incidence vector. Therefore, the loss is piecewise constant with respect to $\widehat{\boldsymbol{C}}_n$, and its gradient is zero almost everywhere and undefined at path-switching boundaries. This non-differentiability prevents direct computation of the gradient of the loss function $\nabla_{\widehat{\boldsymbol{C}}_n} \ell(r_n, \mathrm{SP}(\widehat{\boldsymbol{C}}_n))$.

### 3.3.3 iMLE surrogate loss and gradient

Regarding the backward pass, Eq. (5) denotes the gradient of the loss function with respect to $\boldsymbol{\theta}$, which is decomposed by the chain rule. Because $\widehat{\boldsymbol{C}}_n$ is sampled by the reparameterization trick using the mean and standard deviation returned from the perception model [$f_{\boldsymbol{\theta}}(\boldsymbol{x}_n)$], the Jacobian, $\partial_{\boldsymbol{\theta}} \widehat{\boldsymbol{C}}_n$, is computed directly from the Jacobian, $\partial_{\boldsymbol{\theta}} f_{\boldsymbol{\theta}}(\boldsymbol{x}_n)$. These two Jacobians can be computed using the auto-differentiation scheme provided by any deep-learning frameworks.

$$\nabla_{\boldsymbol{\theta}} L_n(\boldsymbol{\theta}) = \partial_{\boldsymbol{\theta}} \widehat{\boldsymbol{C}}_n{}^{T} \nabla_{\widehat{\boldsymbol{C}}_n} L_n(\boldsymbol{\theta}) \qquad (5)$$

On the other hand, the gradient, $\nabla_{\widehat{\boldsymbol{C}}_n} L_n(\boldsymbol{\theta})$, of the loss function with respect to the perceived link-cost vector $\widehat{\boldsymbol{C}}_n$ cannot be computed directly. As illustrated in Fig. 2, the CO layer maps a continuous cost vector to a discrete route. Consequently, small perturbations of $\widehat{\boldsymbol{C}}_n$ usually do not change the selected route, so the loss remains unchanged over most regions of the cost domain. The loss takes the form of a piecewise constant function, and its gradient is therefore zero almost everywhere except for path-switching points. By contrast, at the path-switching boundaries, the gradient is not defined. This non-differentiability prevents the direct use of a gradient-based algorithm for minimizing the original loss. Therefore, the original loss needs to be replaced by a smooth surrogate whose gradient provides a meaningful descent direction for backpropagation through the shortest path finding layer.

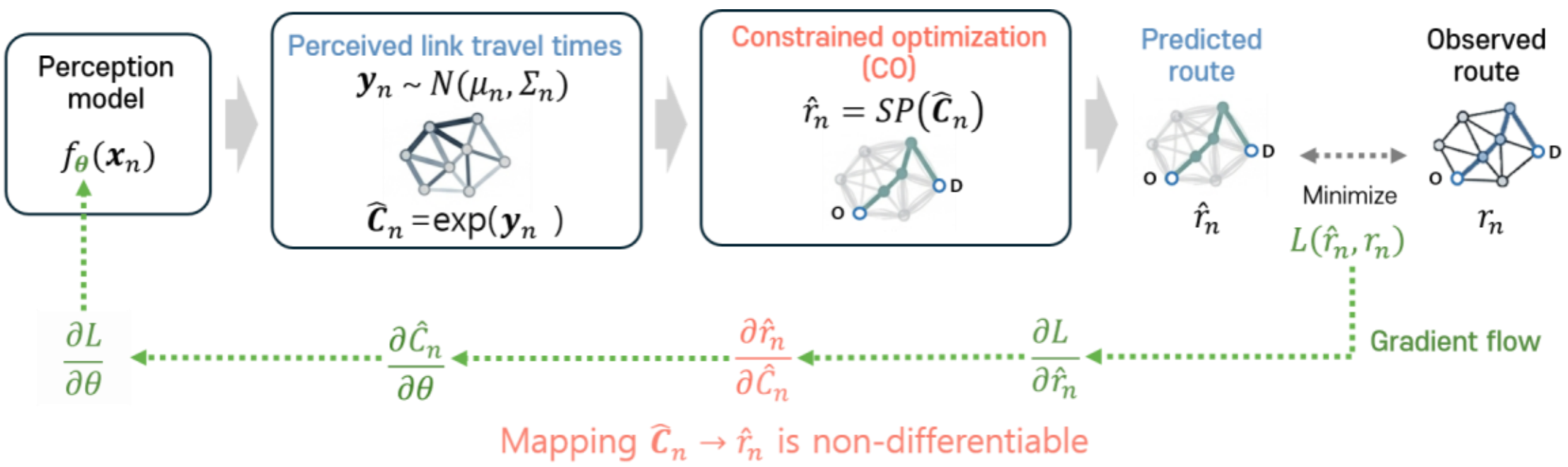

Fig. 2 Non-differentiability of the shortest-path finding layer

Niepert et al. (2021) introduced a surrogate loss $\mathcal{L}_n$, that is defined for differentiation while preserving the learning signal of the original objective ($L_n$). Prior to defining the surrogate loss, a probabilistic mass function (PMF) over routes is defined. Eq. (6) denotes a softmax PMF function that belongs to exponential family. Here, the temperature $\tau$ is a hyperparameter to control the degree of sharpness.

$$P_{\widehat{\boldsymbol{C}}_n}(r_n) = \frac{e^{-\widehat{\boldsymbol{C}}_n{}^T r_n/\tau}}{\sum_{r\in\mathcal{P}_n} e^{-\widehat{\boldsymbol{C}}_n{}^T r/\tau}} \quad (6)$$

Here, $\mathcal{P}_n$ denotes the set of feasible paths connecting the origin and destination of trip $n$. Although this distribution is useful for deriving the iMLE surrogate gradient, it is not evaluated explicitly because enumerating all feasible paths in $\mathcal{P}_n$ is computationally intractable on a large network.

Eq. (7) denotes the gradient of the log-likelihood, $\ln P_{\widehat{\boldsymbol{C}}_n}(r_n)$, with respect to the link cost, $\widehat{\boldsymbol{C}}_n$. The simple derivation of the gradient stems from the exponential family assumption, this gradient will be used later when computing the gradient of the surrogate loss ($\mathcal{L}_n$).

$$\nabla_{\widehat{\boldsymbol{C}}_n} \ln P_{\widehat{\boldsymbol{C}}_n}(r_n) = -r_n + \mathbb{E}_{r_n \sim P_{\widehat{\boldsymbol{C}}_n}(r_n)}[r_n] \quad (7)$$

The surrogate loss, $\mathcal{L}_n$, is defined by averaging the logarithm of the PMF over a chosen target distribution. The PMF of the target distribution is also chosen to be a softmax function, with a different set of link costs ($\widehat{\boldsymbol{C}}'_n$). Eq. (8) defines the surrogate loss to be the negative expectation of log-likelihood over the target distribution [$=P_{\widehat{\boldsymbol{C}}'_n}(r_n)$]. Minimizing the surrogate loss is equivalent to minimizing the KL-divergence between the target and incumbent PMFs, which leads the incumbent PMF as closely to the target PMF as possible.

$$\mathcal{L}_n(\widehat{\boldsymbol{C}}_n, \widehat{\boldsymbol{C}}'_n) = -\mathbb{E}_{r_n \sim P_{\widehat{\boldsymbol{C}}'_n}(r_n)}[\ln P_{\widehat{\boldsymbol{C}}_n}(r_n)] \quad (8)$$

The gradient, $\nabla_{\widehat{\boldsymbol{C}}_n}\mathcal{L}_n(\widehat{\boldsymbol{C}}_n, \widehat{\boldsymbol{C}}'_n)$, leads to Eq. (9) by simple algebra based on Eq. (7), which will substitute for the gradient, $\nabla_{\widehat{\boldsymbol{C}}_n} L_n(\boldsymbol{\theta})$, of the original loss function.

$$\nabla_{\widehat{\boldsymbol{C}}_n}\mathcal{L}_n(\widehat{\boldsymbol{C}}_n, \widehat{\boldsymbol{C}}'_n) = \mathbb{E}_{r_n \sim P_{\widehat{\boldsymbol{C}}'_n}(r_n)}[r_n] - \mathbb{E}_{r_n \sim P_{\widehat{\boldsymbol{C}}_n}(r_n)}[r_n] = \mu(\widehat{\boldsymbol{C}}'_n) - \mu(\widehat{\boldsymbol{C}}_n) \quad (9)$$

Two expectations over all possible routes in Eq. (9) cannot be computed at this stage. Eq. (10) shows that the expectation of routes for a given $\widehat{\boldsymbol{C}}_n$ can be approximated by the average of the MAP state with perturbation in $\widehat{\boldsymbol{C}}_n$. More concretely, if path-level Gumbel perturbations were available, perturb-and-MAP sampling exactly reproduces the exponential family PMF. Since all feasible paths cannot be enumerated, the implementation in Eq. (10) uses structured link-level perturbations and repeated shortest-path evaluations to approximate the required moments. The perturbation-induced distribution is therefore used as a computational device for gradient estimation. Computing the $MAP$ state, equivalent to SP finding operation, is typically much less expensive than computing the expectation [$\mu(\cdot)$] over all possible routes which is almost impossible in practice.

$$\nabla_{\widehat{\boldsymbol{C}}_n}\mathcal{L}_n(\widehat{\boldsymbol{C}}_n,\widehat{\boldsymbol{C}}'_n) = \mu(\widehat{\boldsymbol{C}}'_n) - \mu(\widehat{\boldsymbol{C}}_n) \approx \mathbb{E}_{\boldsymbol{\epsilon}\sim p(\boldsymbol{\epsilon})}\left[MAP(\widehat{\boldsymbol{C}}'_n + \boldsymbol{\epsilon}) - MAP(\widehat{\boldsymbol{C}}_n + \boldsymbol{\epsilon})\right] \quad (10)$$

The gradient, $\nabla_{\widehat{\boldsymbol{C}}_n}\mathcal{L}_n(\widehat{\boldsymbol{C}}_n,\widehat{\boldsymbol{C}}'_n)$, is approximated once again by the simulated version, $\widehat{\nabla}_{\widehat{\boldsymbol{C}}_n}\mathcal{L}_n(\widehat{\boldsymbol{C}}_n,\widehat{\boldsymbol{C}}'_n)$. Eq. (11) denotes the simulated gradient with random draws $\boldsymbol{\epsilon_i}$'s. Note here that the same random draw ($\boldsymbol{\epsilon_i}$) should be used for averaging both $MAP$ states. In practice, the sample size ($s$) can be reduced to 1 if the batch size is sufficiently large. For this case, each element in the $MAP$ difference vector belongs to one of -1, 0 , and +1.

$$\widehat{\nabla}_{\widehat{\boldsymbol{C}}_n}\mathcal{L}_n(\widehat{\boldsymbol{C}}_n,\widehat{\boldsymbol{C}}'_n) = \frac{1}{s}\sum_{i=1}^{s}\left[MAP(\widehat{\boldsymbol{C}}'_n + \boldsymbol{\epsilon_i}) - MAP(\widehat{\boldsymbol{C}}_n + \boldsymbol{\epsilon_i})\right], \text{ with } \boldsymbol{\epsilon_i}\sim p(\boldsymbol{\epsilon}) \quad (11)$$

The gradient of the surrogate loss leads to the difference between two $MAP$ route states when the incumbent and target link cost vectors are given. The next step to compute the gradient is to determine the target link travel cost vector, $\widehat{\boldsymbol{C}}'_n$, which is equivalent to choosing a target distribution. Domke (2010) described that a target distribution can be rooted in perturbation-based implicit differentiation (PID). Eq. (12) denotes the finite difference $[\lambda\nabla_{r_n}\ell(r_n,\hat{r}_n)]$ between two link travel cost vectors.

$$\widehat{\boldsymbol{C}}'_n = \widehat{\boldsymbol{C}}_n - \lambda\nabla_{\hat{r}_n}\ell(r_n,\hat{r}_n) \quad (12)$$

Here, $\lambda > 0$ is a hyperparameter to control the perturbation intensity. The reciprocal $1/\lambda$ incorporated into the global learning rate, $\eta$, for the gradient descent algorithm (Niepert et al., 2021). Eq. (13) shows that the stochastic gradient algorithm to update the model parameter, $\boldsymbol{\theta}$, at every iteration step. Here, $B$ is batch size.

$$\boldsymbol{\theta}' = \boldsymbol{\theta} - \frac{\eta}{B}\sum_{n=1}^{B}\partial_{\boldsymbol{\theta}}\widehat{\boldsymbol{C}}_n{}^{T}\widehat{\nabla}_{\widehat{\boldsymbol{C}}_n}\mathcal{L}_n\left(\widehat{\boldsymbol{C}}_n,\widehat{\boldsymbol{C}}_n - \lambda\nabla_{\hat{r}_n}\ell(r_n,\hat{r}_n)\right) \quad (13)$$

The last step to obtain the gradient is to determine a probabilistic distribution for perturbations ($\boldsymbol{\epsilon}$). Niepert et al. (2021) proved that $\boldsymbol{\epsilon}$ should be drawn from the sum of Gamma (SoG) distribution, so that the Gumbel-Max trick (Papandreou and Yuille, 2010) can reproduce the softmax PMF shown in Eq. (6). According to Theorem 1 in Niepert et al. (2021), perturbing each element of link travel cost vector, $\widehat{\boldsymbol{C}}_n$, by a random draw $\epsilon_i\sim SoG(\kappa,\tau,g)$ leads to perturbing the path cost, $\widehat{\boldsymbol{C}}_n{}^{T}r_n$, with a random draw $\zeta(r_n)\sim Gumbel(0,\tau)$, which results in the Gumbel-max trick (Jang et al., 2017). A $SoG(\kappa,\tau,g)$ distribution has three parameters. $\kappa$ denotes the number of non-zero entries in a route vector to determine both parameters $(1/\kappa,\kappa/i)$ of $Gamma$ PDFs in $SoG$ distribution, $\tau$ is a parameter matching the temperature in the exponential family PMF shown in Eq. (6), and $g$ denotes how many $Gamma$ PDFs are summed in a SoG distribution. Eq. (14) denotes the PDF of $SoG(\kappa,\tau,g)$, where $\kappa$ and $g$ should be a positive integer, and the adopted parameter values are provided in Appendix A.

$$SoG(\kappa,\tau,g) = \frac{\tau}{\kappa}\left[\sum_{i=1}^{g}[Gamma(1/\kappa,\kappa/i)] - \log g\right] \quad (14)$$

The iMLE method derives the approximate gradient of the original loss function with respect to perceived link travel costs ($\widehat{\boldsymbol{C}}_n$). In summary, the method smooths the SP layer and derives the approximate gradient to find a local optimum, preserving learning directions of the original loss function.

Although differentiation through the discrete optimization layer is successful, the proposed framework still faces a more fundamental problem. The adopted supervision to match the estimated and observed routes cannot derive a unique set of perceived link travel costs. Intuitively, the shortest path found on a set of link travel costs does not vary by doubling all link travel costs in the set. This implies that an infinite number of perceived travel cost sets that reproduce a single ground truth route can be found. This scaling issue has not been discussed in previous iMLE studies. Niepert et al. (2021) focused only on differentiating the discrete optimization not mentioning the issue. On the other hand, Mandi et al. (2024) averted the problem by assuming that the ground truth latent variables (=perceived link cost in the present study) could be observed. Because the central objective of the proposed model is to recover behaviorally meaningful perceived link costs, we add a regularization term to stabilize estimation and reduce the scaling challenge. Section 3.3.4 describes how regularizing the distribution of travel costs releases the problem and modifies the loss function to regularize perceived link travel costs.

### 3.3.4 Regularization for cost-scale anchoring

To mitigate the cost-scale ambiguity, we regularize the output returned from the perception model for cost-scale anchoring. The baseline distribution is obtained from observed link travel times. After taking the logarithm to the observed link travel times, the mean ($\tilde{\mu}_a$) and the standard deviation ($\tilde{\sigma}_a$) are computed. The KL-divergence between two Gaussian distributions leads to the closed form that can be easily handled while training the model.

Eq. (15) denotes the reinforced loss function [$\tilde{L}(\boldsymbol{\theta})$] with the KL-divergence [$D_{KL}(\cdot)$], where $\xi$ is a hyperparameter that weighs the regularization with respect to the original loss.

$$\tilde{L}(\boldsymbol{\theta}) = L(\boldsymbol{\theta}) + \xi D_{KL}(\mathcal{N}(\mu_n, \Sigma_n)||\mathcal{N}(\tilde{\boldsymbol{\mu}}, \tilde{\boldsymbol{\Sigma}})) \qquad (15)$$

where $\tilde{\boldsymbol{\Sigma}} = \mathrm{Diag}(\tilde{\boldsymbol{\sigma}}^2)$ is the diagonal matrix containing the observed standard-deviation vector $\tilde{\boldsymbol{\sigma}}$. Here, $(\mu_n, \Sigma_n)$ is the mean and variance for reparameterization, which are returned from the perception function [$f_{\boldsymbol{\theta}}(\boldsymbol{x}_n)$] approximated by a neural net for embedding input features to the perceived travel cost. On the other hand, $\tilde{\boldsymbol{\mu}}$ and $\tilde{\boldsymbol{\Sigma}}$ should be given in advance so that minimizing the KL-divergence renders the perceived link travel cost not to stray away from them. The derivative of the KL-divergence with respect to $\boldsymbol{\theta}$ can be analytically computed.

The overall training procedure, combining perceived-cost sampling, perturb-and-MAP inference, iMLE-based surrogate gradient estimation, and KL regularization, is summarized in Algorithm 1. The values applied to the hyperparameters in the algorithm are listed in Appendix A.

**Algorithm 1. Training algorithm**

**Input:** Training data $\mathcal{D} = \{(\mathbf{x}_n, r_n)\}_{n=1}^{N}$ ; observed link-wise data $(\tilde{\mu}, \tilde{\Sigma})$ ; hyperparameters $\lambda$(perturbation scale), $\xi$(weight for KL-divergence), $s$ (perturbation sample size), $\eta$(Learning rate), and $\kappa$, $\tau$, and $g$ (SoG parameters) (see Appendix A).

**Repeat** until convergence

Sample a mini batch $\mathcal{B} \subset \mathcal{D}$.

**For** each trip $n \in \mathcal{B}$:

**Obtain** the latent perceived-cost distribution parameters:

$$(\mu_n, \Sigma_n) = f_\theta(\mathbf{x}_n)$$

**Sample** a perceived link-cost vector:

$$\boldsymbol{y}_n \sim \mathcal{N}(\mu_n, \Sigma_n),\ \widehat{\boldsymbol{C}}_n = e^{\boldsymbol{y}_n}$$

**Draw** perturbations:

$$\epsilon_i \sim SoG(\kappa, \tau, g),\ i = 1, \dots, s.$$

**Generate** the predicted route through the shortest path / MAP layer:

$$\hat{r}_n = \mathrm{MAP}(\widehat{\boldsymbol{C}}_n + \epsilon_1)$$

**Compute** the route-matching loss:

$$\ell(r_n, \hat{r}_n)$$

**Construct** the target cost vector using the finite-difference perturbation:

$$\widehat{\boldsymbol{C}}'_n = \widehat{\boldsymbol{C}}_n - \lambda \nabla_{\hat{r}_n} \ell(r_n, \hat{r}_n)$$

**Estimate** the surrogate gradient with respect to $\hat{C}_n$:

$$\widehat{\nabla}_{\hat{C}_n} \mathcal{L} = \frac{1}{s} \sum_{i=1}^{s} \left[ \mathrm{MAP}(\widehat{\boldsymbol{C}}'_n + \epsilon_i) - \mathrm{MAP}(\widehat{\boldsymbol{C}}_n + \epsilon_i) \right]$$

**Compute** the KL regularization term:

$$\mathcal{R}_n = D_{KL}\left( \mathcal{N}(\mu_n, \Sigma_n) \parallel \mathcal{N}(\tilde{\mu}, \tilde{\Sigma}) \right)$$

**end for**

**Compute** the mini-batch gradient

$$\mathbf{g}_\theta = \frac{1}{|\mathcal{B}|} \sum_{n \in \mathcal{B}} \left[ (\partial_\theta \widehat{\mathbf{C}}_n)^\top \widehat{\nabla}_{\hat{C}_n} \mathcal{L} + \xi \nabla_\theta R_n \right].$$

**Update** parameters

$$\boldsymbol{\theta}' \leftarrow \boldsymbol{\theta} - \eta \mathbf{g}_\theta.$$

**Return** $\boldsymbol{\theta}$

## 3.4 Architecture of the perception model

Fig. 3 illustrates the architectural design of the perception model, $f_\theta(\mathbf{x}_n)$, which lies in the proposed pipeline, to embed multi-dimensional covariate data, $\mathbf{x}_n$, into the latent distribution, specifically the mean ($\mu_n$) and variance ($\Sigma_n$) of $ln\hat{\mathbf{C}}_n$. The model takes the form of a hierarchical neural network comprising four distinct components. The first three sub-networks function as parallel encoders designed to extract high-level representations from personal information, trip context, and network state, respectively. For this multi-stream encoding, the input covariate vector, $\mathbf{x}_n$, is partitioned into subsets ($\boldsymbol{x}_n^{person}$, $\boldsymbol{x}_n^{trip}$, $\boldsymbol{x}_n^{net}$) tailored for each respective sub-model. The final sub-network maps these three output representations into the parameters of the perceived travel cost distribution. Detailed architecture of the four sub-networks can be found in Appendix B. Furthermore, ablation tests are conducted in Section 5.2, removing each of sub-networks.

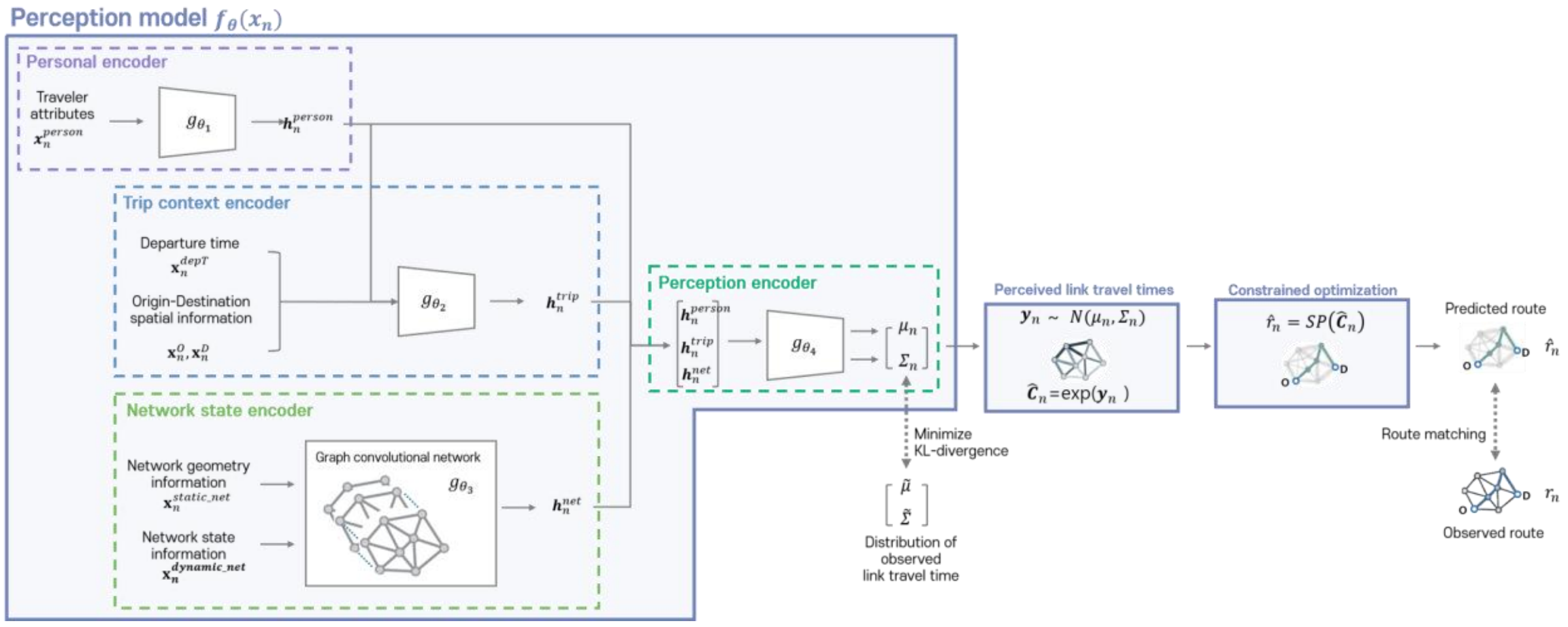

Fig. 3 Architecture of perception model [$f_\theta(\mathbf{x}_n)$]

### 3.4.1 Personal encoder

The personal encoder maps a vector of individual characteristics ($\boldsymbol{x}_n^{person}$) into latent representations ($\boldsymbol{h}_n^{person}$), which are subsequently used to derive the distribution of personalized travel costs. This process is parameterized by a feed-forward neural network as follows:

$$\boldsymbol{h}_n^{person} = g_{\theta_1}(\boldsymbol{x}_n^{person}).$$

where $\theta_1$ denotes the set of trainable parameters within the sub-network. The model's predictive power is maximized by incorporating a comprehensive suite of potential covariates, ranging from socio-demographic attributes such as gender, age, and income to behavioral and attitudinal variables. Furthermore, the inclusion of psychometric propensities, typically collected through stated-preference surveys, can further enhance the model's ability to capture heterogeneous decision-making results. The significance of such personal variables in uncovering the latent motives behind transportation-related choices has been extensively documented in previous studies (Kim et al., 2014; Lee et al., 2012; Park et al., 2011; Sohn and Yun, 2009).

In the absence of explicit socio-demographic data, as the current study relies solely on vehicle trajectory data, we focus on capturing latent driving habits implicitly embedded within individual trajectories. It is hypothesized that a driver's speed profile serves as a proxy for their unique driving habit. For instance, an aggressive driver typically exhibits frequent and abrupt speed changes, whereas a more cautious driver tends to maintain a stable and modest speed throughout the journey. To reflect this, the proposed personal encoder [$g_{\theta_1}(\cdot)$] utilizes the mean and variance of jerks, accelerations (decelerations), and longitudinal and angular speeds observed over a driver's past trajectory as its primary input features. More concretely, for each trip, driving-habit features are computed only from trajectories of the same driver collected before the departure date of the trip.

This methodological choice is strongly supported by prior literature demonstrating that idiosyncratic driver behaviors can be effectively identified from granular vehicle trajectory data (Hallac et al., 2016; Fung et al., 2017). In particular, we adopt the feature construction strategy proposed by Chowdhury et al. (2018), who reported a driver identification accuracy of up to 82.3% using similar kinematic variables. By leveraging these high-fidelity proxies, the model can differentiate between travelers and account for individual-specific perception patterns even without access to traditional demographic information.

### 3.4.2 Trip context encoder

The perceived travel cost is further assumed to be influenced by explicit trip-related attributes, $\boldsymbol{x}_n^{trip} = [\boldsymbol{x}_n^{depT}, \boldsymbol{x}_n^{o}, \boldsymbol{x}_n^{d}]$. The departure time, $\boldsymbol{x}_n^{depT}$, often serves as a proxy for trip purpose; for instance, trips during morning and evening peak hours typically correspond to commuting and returning home, respectively. Such temporal variations represent a unique variable that significantly affects route choice behavior. In addition, the spatial distribution of the origin and destination fundamentally shapes the traveler's choice set. To vectorize these spatial inputs, we simply add 2-D coordinates into the origin- and destination-specific variables ($\boldsymbol{x}_n^{o}$ and $\boldsymbol{x}_n^{d}$). Empirical testing verified that this simple scheme outperforms the rigorous 2-D positional encoding.

To supplement these purely locational features, we further augment $\boldsymbol{x}_n^{o}$ and $\boldsymbol{x}_n^{d}$ by concatenating land-use variables. Specifically, the model incorporates building floor areas, categorized by land-uses, within a 500-meter radius of both the origin and destination. This provides a socio-economic context for a trip's endpoints, which can reflect the nature of the activities being pursued.

The model architecture introduces an additional mechanism to enhance embedding performance, the latent representation of personal attributes is fed into the trip context encoder. Consequently, the latent representation of the trip context ($\boldsymbol{h}_n^{trip}$) becomes a joint function of trip-specific variables ($\boldsymbol{x}_n^{trip}$) and latent variables ($\boldsymbol{h}_n^{person}$) for individual driving habits. The trip context encoder is parameterized by a feed-forward neural network as follows:

$$\boldsymbol{h}_n^{trip} = g_{\theta_2}(\boldsymbol{x}_n^o, \boldsymbol{x}_n^d, \boldsymbol{x}_n^{depT}, \boldsymbol{h}_n^{person})$$

where $\theta_2$ represents the trainable parameters of the network. This hierarchical structure allows the model to capture how the same trip context might be perceived differently depending on the individual's past driving habit.

### 3.4.3 Network state encoder

The network-wide traffic state, $\boldsymbol{x}_n^{net} = [\boldsymbol{x}_n^{static_{net}}, \boldsymbol{x}_n^{dynamic_{net}}]$, is a fundamental determinant of a driver's route choice behavior. This variable is categorized into static and dynamic states to capture both the structural and operational characteristics of the network. The static state, $\boldsymbol{x}_n^{static_net}$, denotes link-level geometric information, including link length, number of lanes, functional road classification, capacity, speed limits, etc. In contrast, the dynamic state, $\boldsymbol{x}_n^{dynamic_net}$, encompasses time-varying operational metrics such as the mean and variance of observed travel times, which are available in the specific time of day for each traveler's departure time. Including the dynamic state as input is rational since every commercial navigation service provides such information on a real time basis.

To effectively capture the spatio-temporal propagation of traffic within the complex urban grid, the network state encoder is parameterized by a graph convolutional network (GCN):

$$\boldsymbol{h}_n^{net} = g_{\theta_3}(\boldsymbol{x}_n^{static_net}, \boldsymbol{x}_n^{dynamic_net})$$

where $\theta_3$ represents the trainable parameters of graph-based architecture. GCNs have been widely adopted to model large-scale urban road networks, proving highly effective for traffic forecasting by accounting for the underlying topology (Liu et al., 2024; Bao et al., 2023; Yu et al., 2020). In Section 5.2, the contribution of adopting GCN is evaluated via ablation test.

In this framework, the network state encoder serves to inform the model of the objective traffic states. This information subsequently becomes baseline from which an individual driver's subjective perception deviates.

### 3.4.4 The final encoder

The three latent representations are concatenated to form the input $[\boldsymbol{h}_n^{person}, \boldsymbol{h}_n^{trip}, \boldsymbol{h}_n^{net}]$ for the final integration encoder, $g_{\theta_4}(\cdot)$. This encoder is parameterized by an additional feed-forward neural network designed to map the consolidated features into the parameters of the perceived travel cost distribution:

$$(\mu_n, \Sigma_n) = g_{\theta_4}(\boldsymbol{h}_n^{person}, \boldsymbol{h}_n^{trip}, \boldsymbol{h}_n^{net})$$

where $\theta_4$ represents the trainable parameters of the final encoder. The derived mean ($\mu_n$) and variance ($\Sigma_n$) are subsequently utilized to sample the perceived travel costs, $\hat{\mathbf{C}}_n$ through the reparameterization scheme (Kingma and Welling, 2014). This trick is critical as it ensures the model pipeline remains differentiable through sampling, thereby enabling the backpropagation of gradients of the loss function during end-to-end training.

Furthermore, the output of the final encoder plays a vital role in regularizing the sampled link travel costs. By ensuring that the perceived travel costs do not deviate excessively from the empirical distribution of observed travel costs, the model maintains the behavioral rationality in interpreting the model results. This regularization mechanism releases the scaling issue of perceived link costs, effectively capturing individual subjective biases from objective traffic states.

# 4. Experimental Set-ups

## 4.1 Testbed and Data Preparation

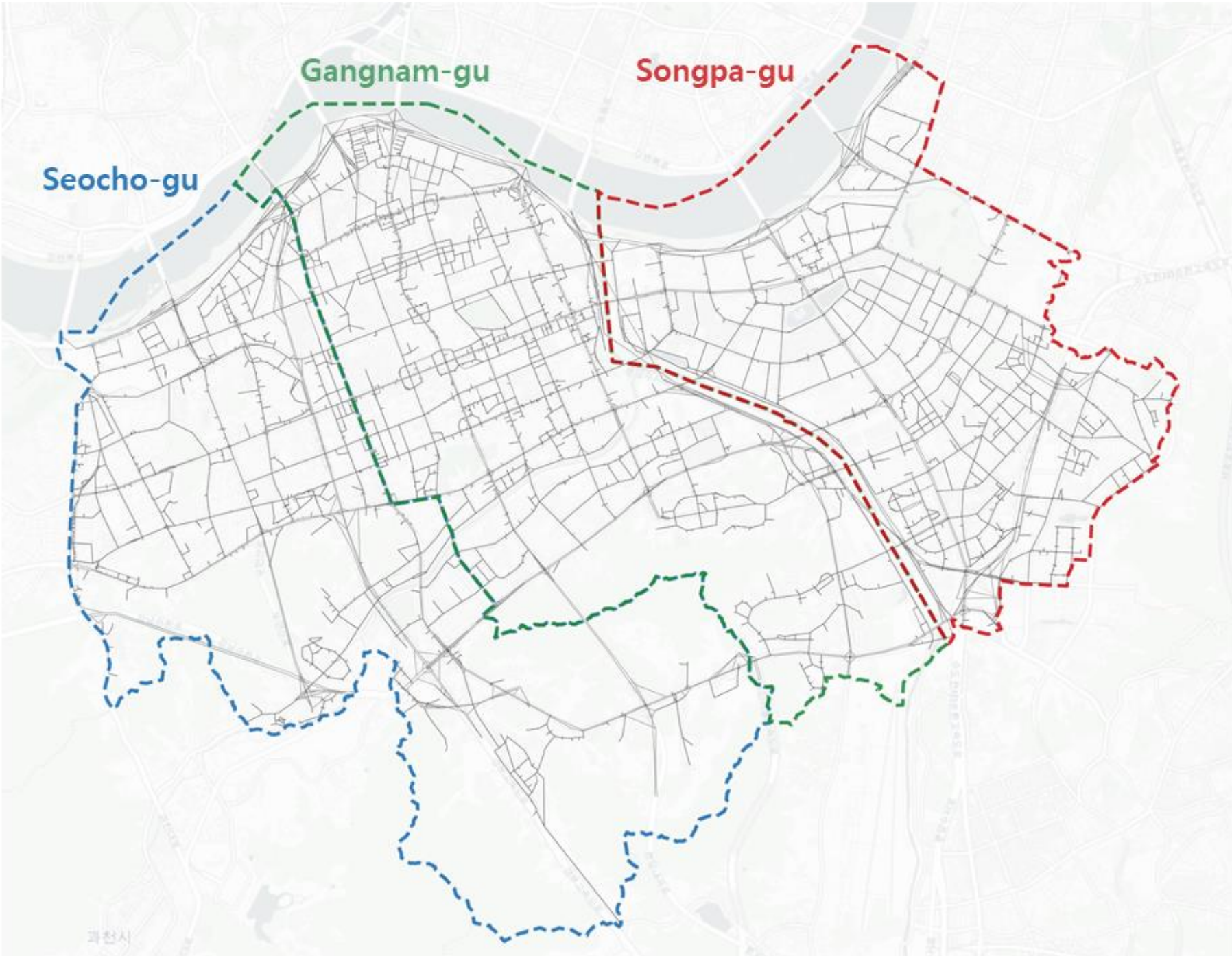

Fig. 4 Road network for testbed in Seoul, Korea

The empirical analysis is conducted on the road network of three major administrative districts in Seoul: Seocho-gu, Gangnam-gu, and Songpa-gu (see Fig. 4). The underlying graph $G = (V, A)$, where $|V|$=7,723 nodes and $|A|$=14,769 directed links. The vehicle trajectory data were provided by a commercial navigation service provider. These trajectories consist of GPS-based position sequences captured from in-vehicle units, which record coordinates from engine startup to shutdown. To transform the raw GPS points into a format suitable for route choice analysis, a map-matching algorithm was employed to snap the trajectories onto the directed links of the road network. To ensure spatial consistency, the dataset was filtered to include only trips where both the origin and destination reside within the study area. The resulting dataset contains 131,792 unique trips. Fig. 5(a) illustrates the trip length distribution, which exhibits a characteristic right-skewed profile. The distribution is characterized by a mode of 2.5 km and a mean of 4.54 km, reflecting the prevalence of relatively short-to-medium distance urban trips within the boundary of study area. Fig. 5(b) illustrates the travel time distribution that takes a similar form with the trip length distribution. The mode is 11 min. and the mean is 18.3 min.

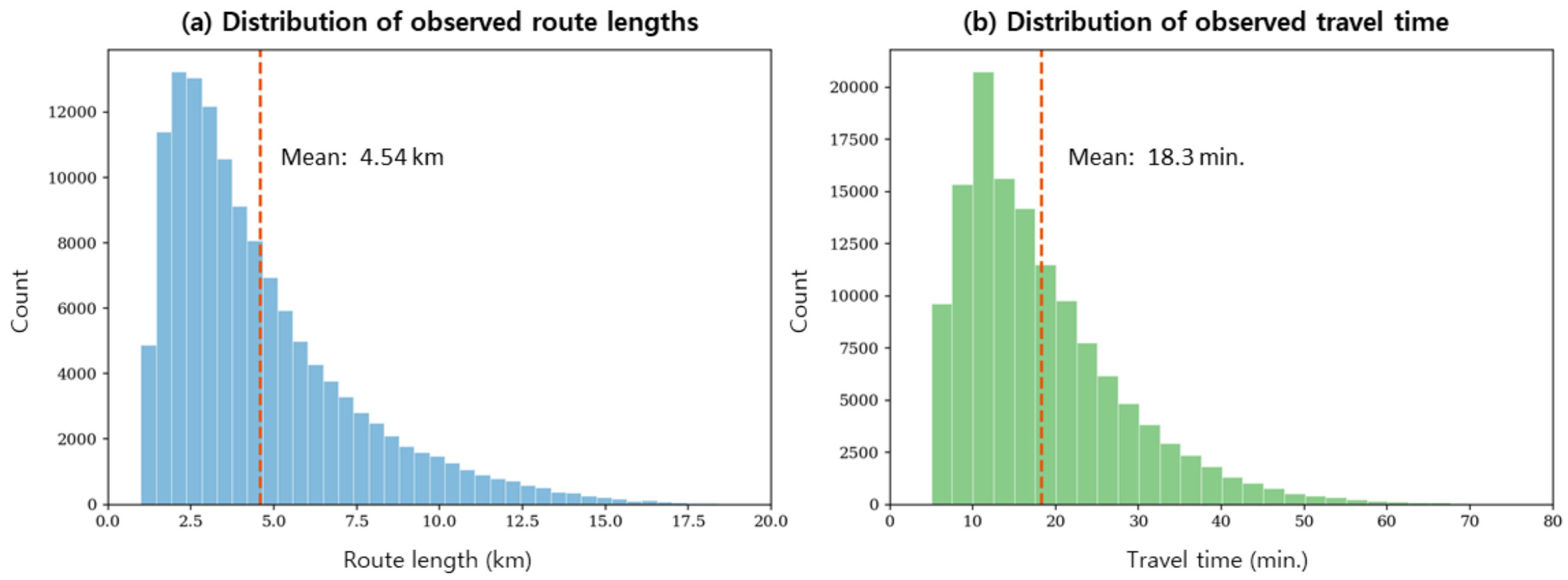

Fig. 5 Distributions of trip length and travel times in the study area

A critical consideration in modeling real-world route choice is the influence of external information on driver behavior. While it is inherently difficult to discern the extent to which any specific observed trajectory was guided by real-time navigation services, drivers are generally presumed to have access to instantaneous traffic data. Consequently, this study adopts the operational assumption that drivers are informed of the ground-truth link travel times at their respective departure times. These observed travel times are integrated into the proposed model as primary inputs, serving a role analogous to the mean travel times employed in conventional discrete choice models. Furthermore, while it is practically impossible for a driver to maintain a perception of travel time on every link within a large-scale urban testbed, our empirical analysis suggests that drivers develop a stable subjective perception of travel times, particularly within the sub-networks encompassing their origin and destination. Detailed evidence supporting this localized perception mechanism is further discussed in Section 5.3.

A driver's speed profile serves as a proxy for individual-specific characteristics. The mean and variance of jerks, accelerations (decelerations), and longitudinal and angular speeds are computed over a trip maker's past trajectory. In this study, single-day trajectories are used for training and testing the proposed modelling framework. Since multi-day trajectories are available for each driver, trajectories observed before the target date to collect data for modelling.

In addition to the vehicle trajectory data, a comprehensive set of covariates was prepared to calibrate and validate the proposed model. For the trip-related attributes, land-use intensity at the trip endpoints was quantified by collecting building floor areas categorized into residential, commercial/office, and recreational uses. These data were aggregated within a 500-meter radius of each origin and destination. Utilizing the GIS-based buffering technique, we processed the spatial data provided by the ministry of land, infrastructure and transport (MOLIT) in SHAPE format to calculate the total floor area for each land-use category within the catchment areas.

The network state data were derived from two primary sources. The static network state, representing the physical infrastructure, was obtained from the official road network database released by the Korea transport database (KTDB). This includes fixed attributes such as link capacity and functional classification. The dynamic network state, capturing real-time traffic conditions, was generated by aggregating individual vehicle trajectories. This synthesized traffic information, encompassing time-varying speeds and travel times, is considered functionally equivalent to the real-time traffic information provided by commercial navigation service providers. These covariate datasets provide the spatio-temporal and socio-economic context influencing each traveler's travel cost perception.

The complete dataset was randomly partitioned into a training set (80%) and a testing set (20%). We constructed a specialized subset to rigorously test how effectively the model reproduces observed paths in scenarios where multiple viable alternative routes exist. This subset focuses on OD pairs characterized by significant route variation, filtered according to a strict three-fold criterion. Specifically, we selected OD pairs that contained at least five observed trips and utilized at least three unique paths. To further ensure that the sampled behavior reflects a high degree of individualistic choice rather than a concentrated flow on a single dominant route, we required the ratio of unique observed routes to the total number of trips to exceed a threshold of 0.9. This process yielded a refined test dataset of 3,668 trips across 546 OD pairs, providing a stringent environment for evaluating the model's capacity to diversify route choice predictions and internalize personalized preferences.

## 4.2 Baseline Models for Comparative Analysis

We compare the proposed framework with five reference models, which are categorized based on two dimensions: formulations and modelling groundwork. Path size logit (PSL) and deep neural net-PSL (DNN-PSL) are classified into path-based models, while recursive logit (ReL) model, adversarial inverse reinforcement learning (AIRL) model, and perturbed utility route choice (PURC) model are link-based. Table 2 lists the reference models according to the corresponding taxonomy. The specification of each model is described in Appendix C to support readers' reproduction attempts.

Table 2. Reference models by route choice formulation and modeling dimensions

| Reference model class | | Formulation dimension | |
|---|---|---|---|
| | | Path-based model | Link-based model |
| Modeling dimension | Discrete choice model | **PSL**<br>(Ben-Akiva and Bierlaire, 1999) | **ReL**<br>(Fosgerau et al., 2013) |
| | Deep learning model | **DNN-PSL**<br>(Marra and Corman, 2021) | **AIRL**<br>(Zhao and Liang, 2023) |
| | Static optimization model | | **PURC** |

(Fosgerau et al., 2022)

PSL is a path-based discrete choice model defined over a finite candidate path set and includes a path-size correction term to account for overlap among alternative routes. DNN-PSL extends PSL by replacing the linear utility specification with a neural-network-based representation, while retaining the same candidate-set-based probabilistic structure. For these path-based reference models, candidate routes were generated for each OD pair using the k-shortest path algorithm with k=10. The link utility is modeled by explanatory variables, and the path utility is obtained by aggregating link-level variables, including travel time, link length, number of lanes, link type, and road class.

The ReL model without path enumeration reformulates route choice as a sequence of link transitions, thereby circumventing the computational burden and potential biases associated with path set generation. It directly utilizes link-level attributes, such as travel time, length, and road class, to estimate transition probabilities. Complementing these discrete choice frameworks, the AIRL-based model treats route choice as a sequential decision-making process within a Markov decision process (MDP) environment. This approach learns context-dependent reward and policy functions directly from observed trajectories. In our implementation of AIRL, the state variables are defined by link-level attributes, while departure time and the remaining distance to the destination are incorporated as context variables to capture the dynamic and goal-directed nature of the observed choices.

PURC model is the latest model free from enumerating alternative paths. The model belongs to static inverse optimization. Link travel time, link length, number of lanes, and road class are chosen as input for PURC. For each OD pair, PURC retrieve link flows for given demand. The individual trajectory is used to derive the link flows in the present study.

The proposed framework is to reproduce an individual route, but baseline models are all choice models or inverse optimization models. To evaluate the path reproduction performance, output of baseline models need to be go through an additional post process.

Unlike the proposed framework, PSL and DNN-PSL generate a probability distribution over a finite candidate set. To ensure a consistent and robust comparison with the observed ground-truth trajectories, we adopt an expectation-based averaging scheme rather than simple stochastic sampling. For each trip, the performance metric is computed between the ground-truth route and every individual route within the candidate set (comprising the top-10 shortest paths). These scores are then aggregated into a representative expected value, weighted by the model-estimated choice probabilities. It is important to note that the candidate set may not always encompass the actual observed route; therefore, the resulting metrics should be interpreted as candidate-set-level performance measures rather than direct, deterministic route-level prediction accuracies.

To ensure a consistent and robust comparison with link-based models, a sampling-based best-matching scheme is devised. For each OD pair, thirty complete routes are generated through repeated stochastic rollouts, and each ground-truth route is matched with the sampled route that yields the highest overlap level. This procedure is deliberately favorable to stochastic link-based baselines because it gives credit when their sampled route set contains a route close to the observed trajectory. Therefore, the resulting scores should be interpreted as upper-bound performance.

Since PURC model cannot return deterministic path output, a conceptual comparison is necessary based on sampling. That is, the estimated optimal link-flow vector is first transformed into transition probabilities by normalizing positive outgoing flows at each node. Thirty complete routes are then sampled through the same random-walk procedure, and the most frequently generated route is selected as the PURC's path prediction. This modal sampled route is compared with the ground-truth route using the same evaluation metrics. This treatment converts PURC's link-flow output into a route-level prediction, allowing it to be compared using the same route-reproduction metrics as the proposed framework.

## 4.3 Evaluation Metrics

To benchmark route reproduction performance, we evaluate the similarity between the predicted and observed routes using three route-level measures. IoU is used as the primary metric, and trip length by links accuracy (TLLA) and edit distance (ED) are reported as supplementary metrics. TLLA supplements IOU by considering the length of overlapping links (Qiu et al., 2024), and ED by reflecting the order-sensitivity in route dissimilarity (Choi et al., 2021; Zhao and Liang, 2023; Qiu et al., 2024). Consequently, these measures capture (i) overlap in traversed road segments, (ii) similarity in total traveled distance, and (iii) order-aware similarity in the link sequence. Unless stated otherwise, the measures are reported as aggregate scores averaged over trips in the test set. Higher values indicate better performance for IoU and TLLA, whereas lower values indicate better performance for ED.

**Intersection-over-Union on links (IoU)**

IoU measures set-level overlap between the predicted and observed link sets. Using the multi-hot link incidence vectors $\hat{r}_n, r_n \in \{0,1\}^{|A|}$, IoU is defined as

$$\mathrm{IoU}(\hat{r}_n, r_n) = \frac{|\hat{r}_n \cap r_n|}{|\hat{r}_n \cup r_n|} = \frac{\hat{r}_n^{\top} r_n}{\| \hat{r}_n \|_1 + \| r_n \|_1 - \hat{r}_n^{\top} r_n} .$$

IoU ranges from 0 to 1, where 1 indicates perfect overlap in observed links. Let $r_n$ and $\hat{r}_n$ denote the observed and predicted route, respectively.

**Trip Length by Links Accuracy (TLLA)**

TLLA is chosen to measure the proportion of the observed route length that is correctly covered by the predicted route. Let $l(e)$ denote the length of link $e$. TLLA for a trip $n$ is defined as

$$\mathrm{TLLA}_n = \frac{\sum_{e \in \hat{r}_n \cap r_n} l(e)}{\sum_{e \in r_n} l(e)}.$$

A higher value indicates better route reproduction.

**Edit Distance (ED) on link sequences**

Normalized edit distance is selected as a supplementing performance index to measure the order-sensitive dissimilarity between the observed and predicted link sequences. Let the observed and predicted routes for trip $n$ be $\pi_n = (a_1, \dots, a_{L_n})$ and $\hat{\pi}_n = (\hat{a}_1, \dots, \hat{a}_{\hat{L}_n})$, respectively. The Levenshtein edit distance $d_{\mathrm{ED}}(\hat{\pi}_n, \pi_n)$ is first necessary to compute this index. $d_{\mathrm{ED}}(\hat{\pi}_n, \pi_n)$ is defined as the minimum number of insertions, deletions, and substitutions required to transform $\hat{\pi}_n$ into $\pi_n$. To make the score comparable across trips of different lengths, the normalized edit distance is computed as follows:

$$\mathrm{ED}_n = \frac{d_{\mathrm{ED}}(\hat{\pi}_n, \pi_n)}{\max(\hat{L}_n, L_n)}.$$

A lower value indicates better route reproduction.

## 5. Results and Discussions

### 5.1 Route Matching Performance and Comparative Analysis

Table 3 presents the comparative performance of the proposed framework against the four reference models on the test dataset. In this evaluation, the IoU is prioritized as the primary performance metric, as it provides a direct spatial measurement of link-level overlaps between the predicted and observed trajectories. To provide a multi-dimensional assessment, the TLLA and ED are incorporated as supplementary metrics, capturing route-length similarity and order-sensitive sequence consistency, respectively.

As reported in Table 3, the proposed framework consistently outperforms all reference models across the three metrics, achieving an IoU of 0.666, a TLLA of 0.926, and an ED of 0.302. Among the baselines, DNN-PSL exhibits the strongest performance within the path-based category, while AIRL emerges as the most competitive link-based alternative. It is important to contextualize these results: the path-based baselines (PSL and DNN-PSL) inherently benefit from an evaluation conducted on a pre-constrained candidate route set, whereas our model performs a direct, unconstrained prediction on the entire network. Despite this difference, the proposed model's superior scores suggest a more faithful reconstruction of both the traversed link sequences and the overall spatial structure of observed routes. The performance of PURC, which is a static optimization model without path enumeration, lies between path- and link-based models. For TLLA, Purc is competitive with the proposed model.

These findings indicate that the decision-focused learning architecture, by internalizing latent perceived travel times, captures the underlying route choice process more effectively than conventional discrete choice, reinforcement learning, and static optimization approaches. The high TLLA and low ED scores further validate that the proposed model does not merely match isolated links but successfully preserves the sequential and topological integrity of the travelers' choices.

Table 3. Route reproduction performance on the test set.

| Models | | Metrics | | |
|---|---|---|---|---|
| | | **IoU** | **TLLA** | **ED** |
| **Proposed model** | | **0.666** | **0.926** | **0.302** |
| **Path-based model** | PSL | 0.527 | 0.631 | 0.385 |
| | DNN-PSL | 0.533 | 0.633 | 0.381 |
| **Link-based model** | ReL | 0.422 | 0.598 | 0.541 |
| | AIRL | 0.445 | 0.573 | 0.576 |
| | PURC | 0.489 | 0.901 | 0.476 |

To further scrutinize the model's capability in capturing behavioral diversity, Table 4 reports the performance metrics specifically for the diversity test set. This subset, characterized by OD pairs with multiple observed route alternatives, serves as a rigorous benchmark to evaluate whether individual-specific perceived travel times can effectively distinguish heterogeneous route-choice patterns. As shown in Table 4, the superiority of the proposed framework becomes even more pronounced in the more complex choice environment. The model achieves an IoU of 0.835, a TLLA of 0.939, and an ED of 0.153, consistently outperforming all competing methods. While the performance of the reference models also improves when restricted to this specialized subset, the magnitude of improvement for the proposed model significantly exceeds that of the baselines. Among reference models, PURC is enhanced the most when using the special subset, and for TLLA the performance of PUCR reaches that of the proposed model. DNN-PSL and AIRL remain the most competitive path-based and link-based models, respectively.

These results provide strong empirical evidence for the study's central hypothesis: when multiple routes are utilized for the same OD pair, characterizing latent, individual-specific perceived travel times allows for a more granular differentiation of route choices than is possible with existing modeling paradigms. Furthermore, the findings suggest that the proposed model does not merely converge toward an aggregate "average" route pattern. Instead, by internalizing the unique perception of each traveler through a decision-focused learning architecture, the model potentially accounts for the underlying heterogeneity that drives divergent route-choice behaviors under identical network conditions (further qualitative interpretations are provided in Sections 5.3 and 5.4).

Table 4. Route reproduction performance on the specialized diversity subset of test data.

| Models | | Metrics | | |
|---|---|---|---|---|
| | | IoU | TLLA | ED |
| **Proposed model** | | **0.835** | **0.939** | **0.153** |
| **Path-based model** | PSL | 0.718 | 0.790 | 0.213 |
| | DNN-PSL | 0.729 | 0.803 | 0.205 |
| **Link-based model** | ReL | 0.482 | 0.705 | 0.480 |
| | AIRL | 0.509 | 0.694 | 0.392 |
| | PURC | 0.768 | 0.929 | 0.215 |

To more rigorously evaluate the stability of the proposed framework, its predictive performance was analyzed across four distinct dimensions: the number of links per path, network path distance, path travel time, and Euclidean OD distance. Each dimension was categorized into three operational levels (low, medium, and high), and the three performance metrics (IoU, TLLA, and ED) were evaluated for each corresponding subset of the test data. Fig. 6 presents the comparative results between the proposed model and the three baseline frameworks.

The proposed model consistently outperforms all baselines across every level and dimension. While it is naturally expected that route reproduction performance might degrade as path complexity and length increase, our model demonstrates a unique and resilient trend. Specifically, the performance gap between the proposed model and the baselines widens as the path length increases.

A particularly noteworthy finding is observed in the TLLA metric, which assesses the degree of spatial overlap by considering the relative length of overlapping segments, rather than merely counting overlapping link for IoU. Unlike link- and path-based baseline models, whose performance significantly deteriorates as paths lengthen, the performance of PURC and the proposed model, as measured by TLLA, remains unchanged. The proposed model's performance even slightly enhances with increased path length. When a path is long and complex at a high level, PURC is slightly inferior to the proposed model. This suggests that the decision-focused learning architecture is exceptionally effective at preserving the integrity of longer routes within complex urban networks. This ability to reliably reproduce long-distance trajectories constitutes a significant advantage of the proposed framework over existing route choice models.

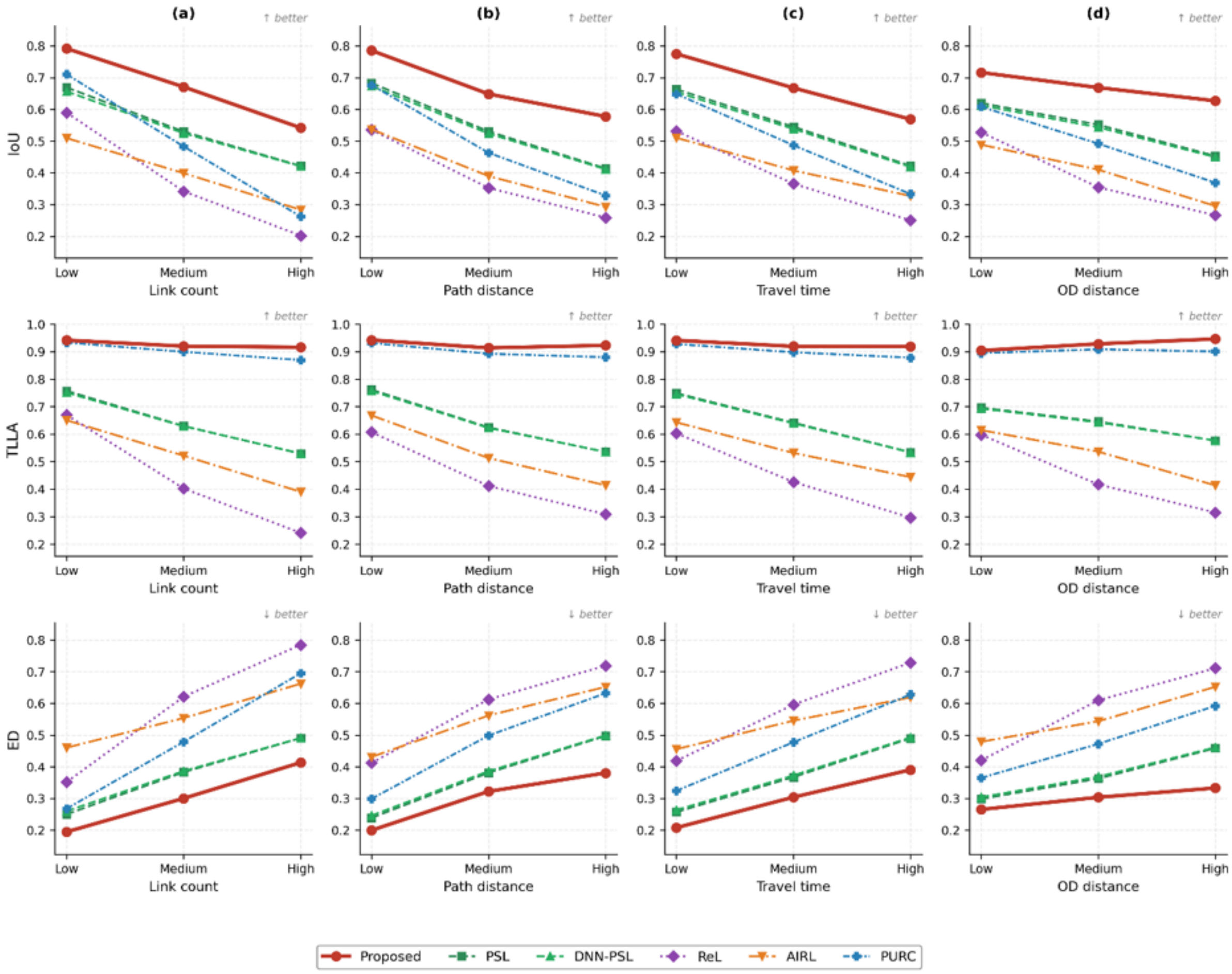

Fig. 6 The model performance comparison with respect to different dimensions

## 5.2 Ablation Analysis

To isolate the contribution of each architectural component in the perception model, an ablation analysis was conducted by systematically removing or replacing the sub-networks described in Section 3.4. Three ablation variants were constructed removing the personal, trip-context, and network state sub-networks. To examine the impact of network encoder in more detail, two more ablation settings are constructed, one removing network state features and the other replacing the GCN architecture with simple dense layers. Finally, the model without KL regularization that anchors the latent perceived cost distribution to the empirical distribution of observed travel costs.

Table 5. Route reproduction performance for ablation models on the test set

| **Models** | **Metrics** | | |
|---|---|---|---|
| | **IoU** | **TLLA** | **ED** |
| **Proposed model** | **0.666** | **0.926** | **0.302** |
| Model without personal encoder | 0.660 | 0.923 | 0.331 |
| Model without trip-context encoder | 0.658 | 0.921 | 0.308 |
| Model without network state features | 0.641 | 0.925 | 0.325 |
| Model replacing GCN with dense layers | 0.578 | 0.919 | 0.382 |
| Model without KL regularization | 0.505 | 0.910 | 0.481 |

Tables 5 summarize the ablation test results. Overall, the largest performance degradation is observed when the KL regularization term is removed. The second most influential component is the GCN-based network encoder. Replacing the graph-based architecture with simple feed-forward neural layers leads to a clear decline in route reproduction performance. This suggests that GCN plays a role of preserving the network topology.

By contrast, removing the personal encoder, trip-context encoder, or network state features causes relatively moderate degradation when testing against whole test data. However, the impact of three encoders becomes more pronounced in the more complex path environment. Table 6 lists the performance of variant models when using the specialized diversity subset of test data. In this setting, the performance degradation of the three input-removal variants is larger than that when full test data are used, suggesting that these contextual inputs become more useful when route choice ambiguity increases. This tendency is more clearly illustrated in Fig. 7. At low complexity levels, the input-removal variants perform similarly to the full model, but the gap gradually widens as route complexity increases. This suggests that individual-specific, trip-specific, and network-state information plays a complementary role that becomes increasingly important when the model must resolve more ambiguous and complex route choice situations.

Table 6. Route reproduction performance on the specialized diversity subset of test data

| **Models** | **Metrics** | | |
|---|---|---|---|
| | **IoU** | **TLLA** | **ED** |
| **Proposed** | **0.835** | **0.939** | **0.153** |
| Without personal encoder | 0.801 | 0.931 | 0.184 |
| Without trip-context encoder | 0.803 | 0.931 | 0.183 |
| Without network state features | 0.803 | 0.930 | 0.183 |
| Without GCN | 0.756 | 0.928 | 0.222 |
| Without KL regularization | 0.531 | 0.838 | 0.492 |

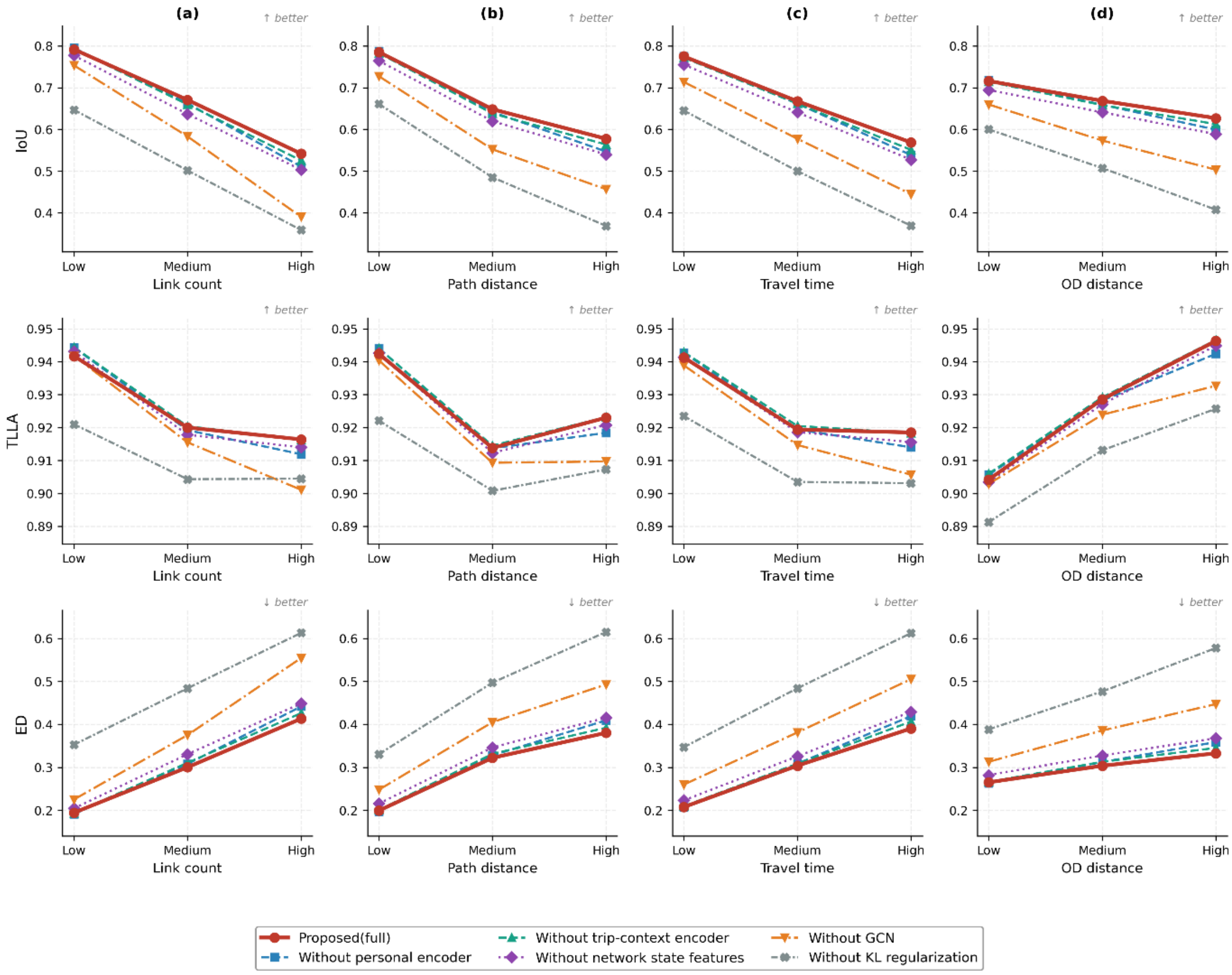

Fig. 7 The model performance comparison for ablation with respect to different dimensions

### 5.3 Interpretation of Perceived Travel Times at Network Level

The most significant advantage of the proposed framework lies in its ability to externalize the latent perceived travel times within a driver's decision-making process, thereby providing an interpretation of individual route-choice motives. Once properly calibrated, the model estimates personalized perceived link travel times within the testbed. However, the assumption that a driver maintains a unique and differentiated perception of travel times across the entire large-scale network is practically unrealistic.

Our empirical findings suggest that the learning process naturally regularizes a driver's perception boundary, narrowing it down to a specific sub-area, which typically covers a corridor-like region encompassing the origin and destination. In areas remote from the chosen route, drivers do not deviate significantly from the baseline travel times provided by navigation services. This implies that travelers do not actively evaluate travel times for links irrelevant to their trip context, aligning with the principle of bounded rationality in spatial cognition.

To delineate this "perception area" and quantify the degree of personalization, we introduce a discrepancy index. This index measures the deviation of the perceived travel time ($\hat{C}_{a,n}$) from the observed mean travel time ($\tilde{\mu}_a$) provided as the model input. For traveler $n$ and link $a$, the time-gap index $\delta_{a,n}$ is defined as follows:

$$\delta_{a,n} = \frac{|\ \hat{C}_{a,n} - \tilde{\mu}_a\ |}{L_a},$$

where $L_a$ denotes the length of link $a$ for scaling the gap index. A higher value of $\delta_{a,n}$ signifies a substantial deviation from the objective traffic state, indicating a strong personalized perception of that specific link. Conversely, an index value near zero suggests that the driver's perception coincides with the observed travel time. This further suggests that the model captures a form of cognitive economy, wherein personalized cost evaluations are not internalized for network segments that fall outside the drivers' immediate decision-making horizon.

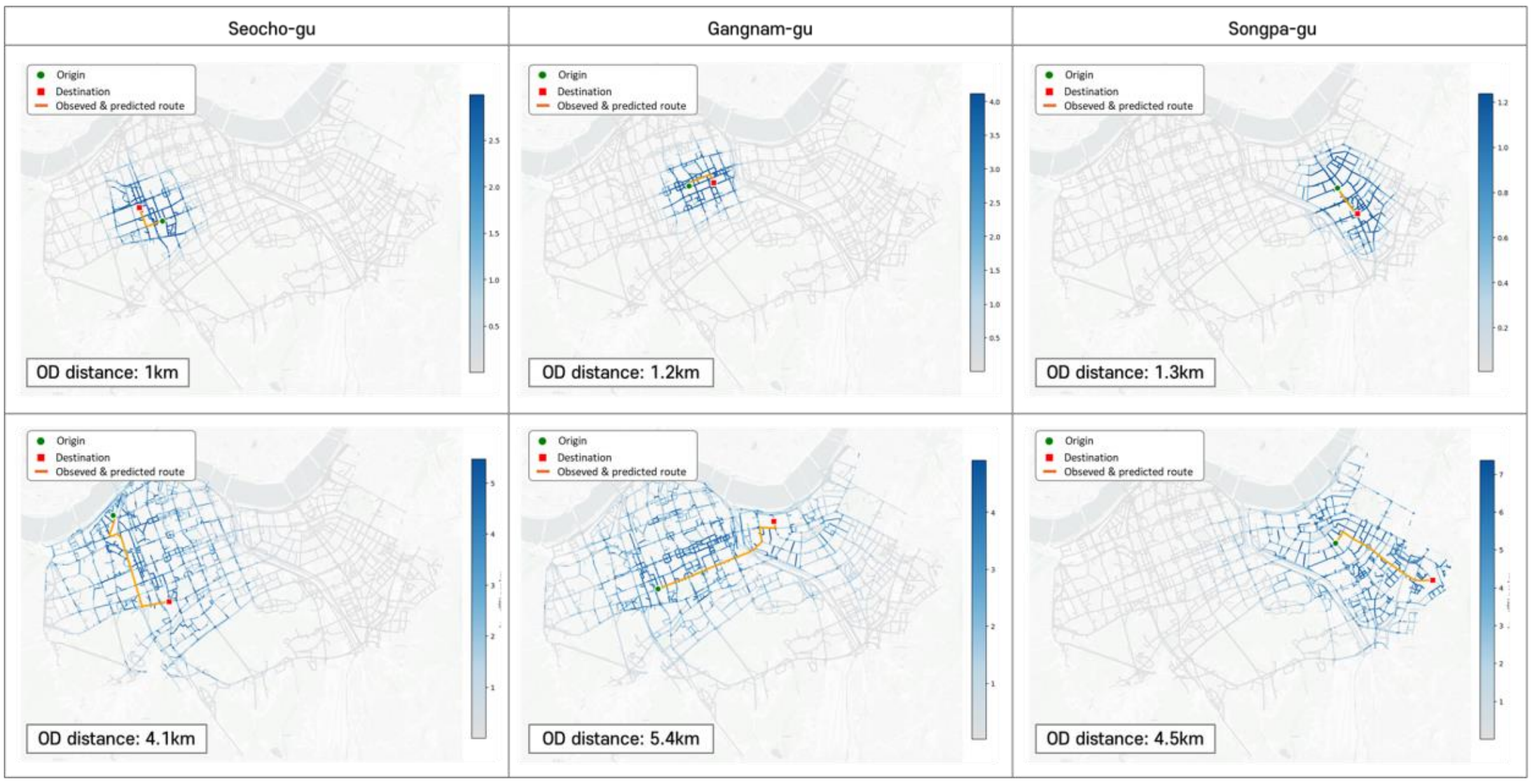

Fig. 8 Spatial distribution of time gaps ($\delta_{a,n}$) for several chosen trips with different OD pairs

Fig. 8 illustrates the spatial distribution of the time-gap index ($\delta_{a,n}$) for six representative trips where the predicted and observed routes are identical. The perception area, highlighted in intense blue, appears to be spatially proportional to the Euclidean distance between the origin and destination. In all depicted cases, significant discrepancies ($\delta_{a,n} > 0$) are concentrated in the vicinity of the origin, the destination, and the connecting corridor. Conversely, links situated further from this OD corridor consistently exhibit values near zero, indicating a convergence to the baseline mean travel times.

As the Euclidean OD distance increases from approximately 1.0–1.3 km to 4.1–5.4 km, the perception region is correspondingly expanded. This contraction and extension of the perception area are behaviorally plausible, reflecting that travelers scale their cognitive efforts according to trip length. Notably, this spatial pattern emerged autonomously through large-scale route data training, without the imposition of any explicit constraints or spatial priors. This suggests that the model effectively internalizes the realistic behavioral tendency of travelers to prioritize information within a localized decision-making corridor rather than considering the entire road network.

This relationship is further quantified in Fig. 9, which plots the radius of the perception area which is defined as the maximum distance encompassing all links with $\delta_{a,n} > 0$ against the Euclidean OD distance. The results indicate a general proportionality between the trip distance and the spatial extent of the perception area. Furthermore, the variance in the radius of perception area increases as the origin and destination become more remote. This increased dispersion suggests that for longer trips, individual-specific preferences and route-choice strategies become more diverse, leading to more heterogeneous spatial configurations of the latent perception area.

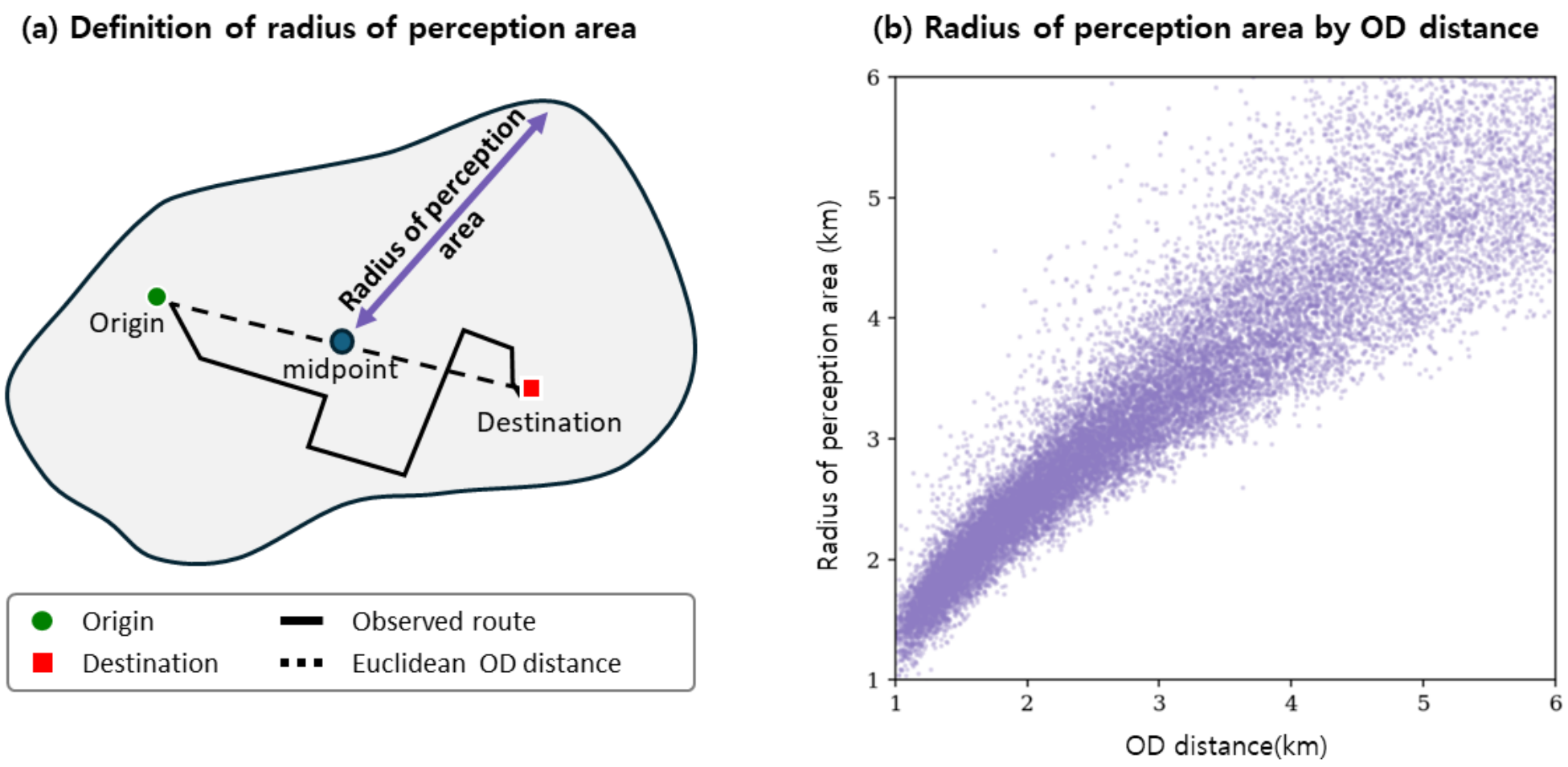

Fig. 9 Relationship between arial OD distance and the radius of perception area

Fig. 10(a) illustrates the link-level variations in estimated perceived travel times ($\widehat{\boldsymbol{C}}_{a,n}$). Links highlighted in intense blue represent segments where travelers' perceptions deviate significantly from one another. Notably, freeways exhibit higher levels of perceptual variance, suggesting that drivers perceive travel times on high-capacity roads in more diverse manners. In Fig. 10(a), freeway links are delineated in yellow lines for readers' better understanding. While some drivers may perceive their travel time to be shorter than the mean perceived travel time, others may internalize a significantly higher perceived travel time for the same segment. It is important to note that this finding is independent of other out-of-pocket costs, as no tolls are charged within the study area. This qualitative observation is further supported by the box plots in Fig. 10(c), which indicate that the interquartile range (IQR) of standard deviations for perceived travel times is substantially larger for freeways than for arterials. Beyond this distinction by road type, certain arterial segments also exhibit high levels of perceptual diversity when compared to other arterial segments, implying that specific road geometries or environmental factors may further contribute to the heterogeneity in driver perception.

In contrast, Fig. 10(b) depicts the variations in observed link travel times recorded throughout the day. There is no distinct evidence to suggest that the variance in observed travel times differs significantly between freeways and arterials. Collectively, Figs. 10(c) and 10(d) show that fluctuations across links in the standard deviation of perceived travel times are, in general, less pronounced than those in objective traffic states. This implies that the variance in actual travel times more fluctuates across links than that in perceived travel times. In particular for arterials, the IQR of standard deviations of perceived travel times is less than that of actual travel times. Although actual travel times are affected by varying traffic conditions across links, travelers tend to less consider these variations when perceiving their own travel times in complex urban road networks.

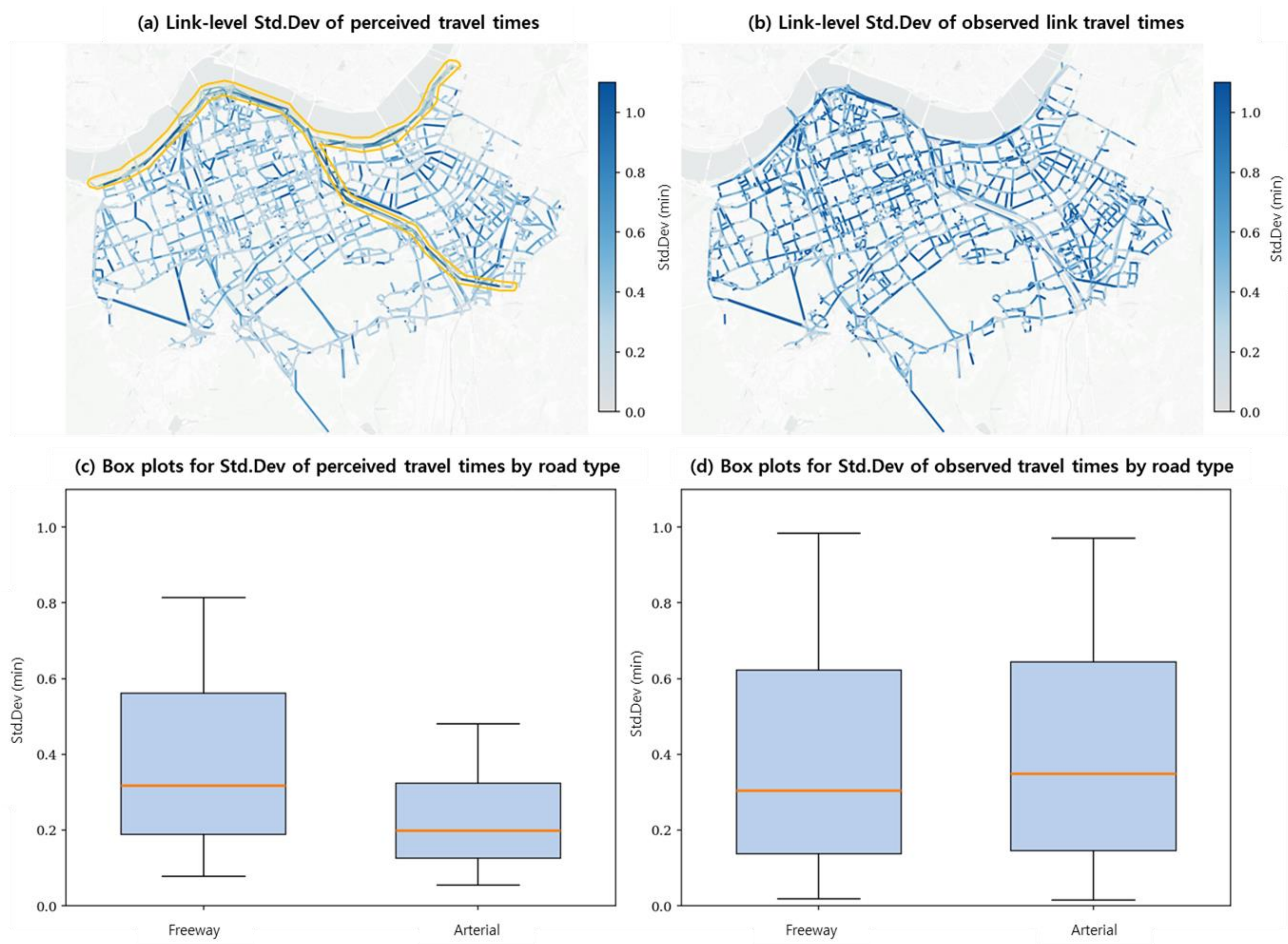

Fig. 10 Variations in perceived and observed link travel times

## 5.4 Interpretation of Individual-Specific Route Choice Behaviors

Beyond network-level interpretations, the proposed model provides a way to interpret the latent motives underlying individual-specific route choice behaviors. The following case study demonstrates why observed routes often deviate from the objective shortest path and how travelers with identical OD pairs and departure times make different routing decisions.

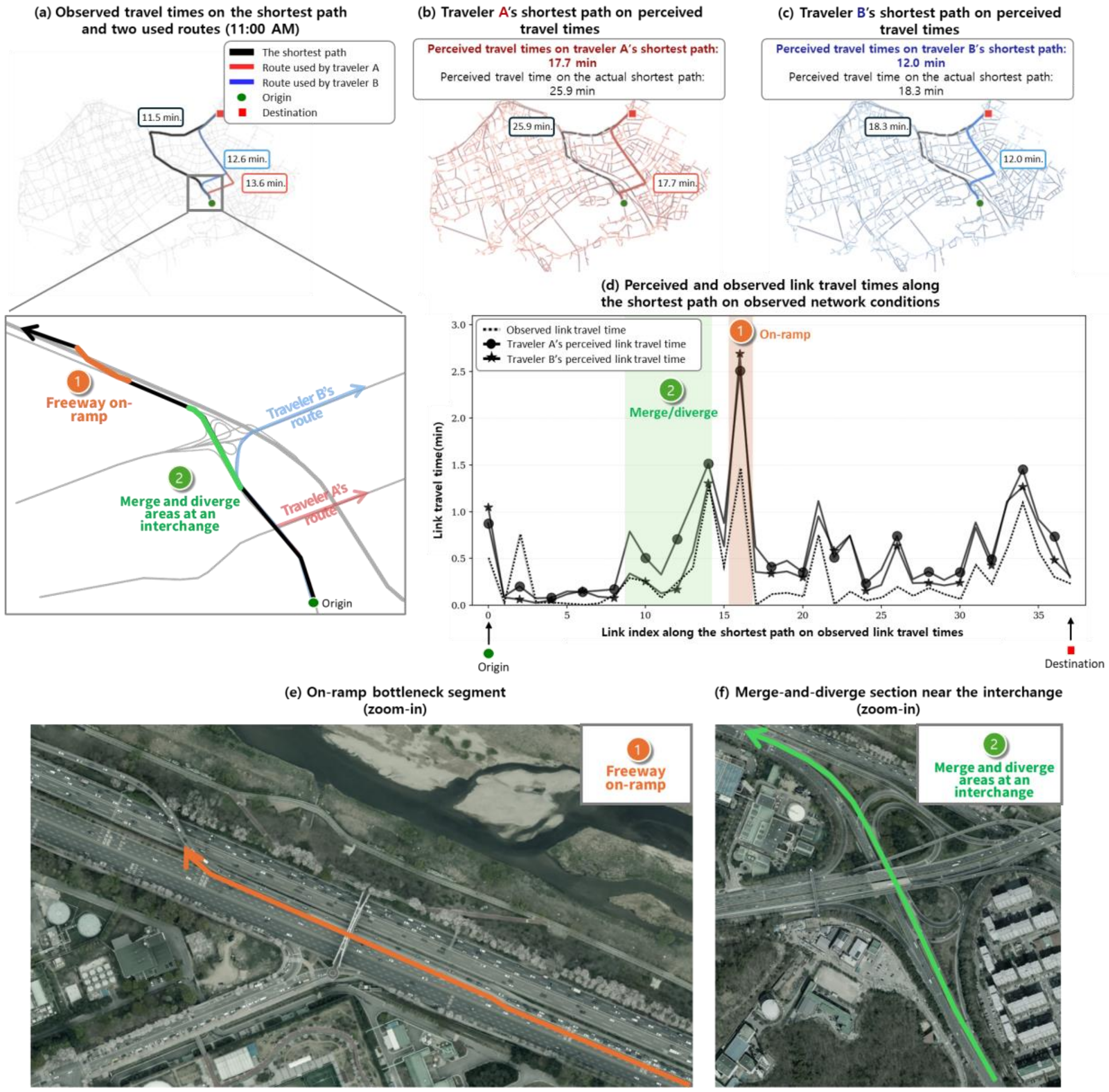

Fig. 11 Explaining differences in individual route choices based on perceived travel times

Fig. 11 illustrates a scenario involving two travelers, A and B, sharing the same OD pair at 11:00 AM. As shown in Fig. 11(a), the objective shortest path under observed network conditions requires 11.5 minutes. However, traveler A selected a route with a travel time of 13.6 minutes, while traveler B chose a route of 12.6 minutes. Despite the objective efficiency of the 11.5-minute path, neither traveler selected it.

Figs. 11(b) and 11(c) display the estimated perceived link travel times for travelers A and B, respectively. For traveler A, the chosen route coincides perfectly with his/her subjective shortest path based on perceived travel times, even though he/she overestimates both travel times on the shortest and used routes. Specifically, traveler A perceives the chosen route to take 17.7 minutes, while estimating the objective shortest path at 25.9 minutes. A symmetrical pattern is observed for traveler B in Fig. 11(c), who underestimates the cost of the chosen route (12.0 min) while perceiving the objective shortest path as significantly more burdensome (18.3 min). These results underscore that travelers do not necessarily follow the objective traffic state; rather, they minimize travel times relative to their own idiosyncratic perception.

The mechanisms driving these deviations are further elucidated in Figs. 11(d)–(f). Fig. 11(d) plots the link-level travel times along the objective shortest path, where the observed travel times (baseline) are contrasted with the subjective perceptions of both travelers. A critical finding is that perceptual deviations are not uniform but are concentrated at specific road segments.

In Section ① (shaded in orange), both travelers perceive travel times to be substantially longer than the true congestion levels. As shown in the satellite imagery in Fig. 11(e), this segment corresponds to a freeway on-ramp prone to recurring traffic jams. The heightened uncertainty associated with merging maneuvers at this location likely causes both travelers to exaggerate the perceived delay, prompting them to avoid the objective shortest path altogether.

The differentiation between travelers A and B is primarily explained by Section ② (shaded in green), which corresponds to a weaving area near an interchange [Fig. 11(f)]. Traveler A exhibits a substantially higher perceived cost in this section, leading him/her to deviate from the objective shortest path to avoid the augmented complexities of the weaving area. In contrast, traveler B perceives Section ② as relatively endurable but chooses an earlier detour to avoid the anticipated burden of the on-ramp in Section ①.

These examples imply that behavioral heterogeneity in route choice stems from divergent individual perceptions of personalized network states. By capturing how travelers uniquely internalize the impedance of complex road geometries such as on-ramps and weaving sections, the proposed model successfully explains why individuals ignore the objective shortest path and why their chosen alternatives differ from one another even under identical external conditions.

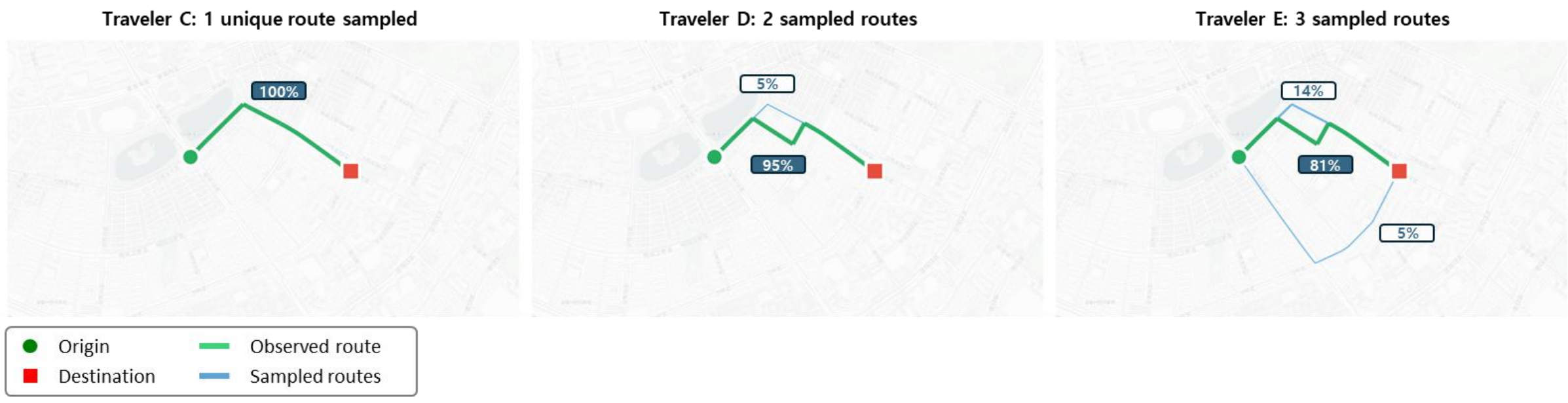

Fig. 12 Monte Carlo simulation of perceived travel times an OD pair

The proposed framework enables the sampling of latent perceived travel times through the encoder $[f_\theta(\mathbf{x}_n)]$ within the model pipeline. To examine the stochastic nature of individual route choice, we selected for a Monte Carlo simulation three travelers sharing an identical OD pair and departure hour. Fig. 12 illustrates the results, where the observed routes are highlighted in green and the additional paths generated through 100 simulations, each case based on a unique set of sampled perceived travel times, are shown in blue. Despite conducting 100 simulations, the number of generated routes ranges from 1 to 3. The labels indicate the selection frequency of each route as the shortest path under the simulated perception of travel times.

The simulation results reveal distinct levels of route adherence among the three travelers. Traveler C exhibits the most stable behavior; all 100 sampled instances consistently converge to the observed route, which also happens to be identical to the objective shortest path. This suggests that traveler C has a highly deterministic perception of travel times with negligible fluctuation. In contrast, travelers D and E, who share the same observed route, demonstrate greater flexibility. Traveler D's observed route remains dominant (95%), but an alternative path emerges in 5% of the simulations. Traveler E displays the highest perceptual variance, with the observed route reproduces in 81% of cases, while two additional alternatives are sampled at frequencies of 14% and 5%. These findings imply that traveler E possesses lower confidence or higher stochasticity in travel time perception compared to the others.

A key contribution of the proposed model is its ability to distinguish between intra-individual heterogeneity (variance within a single driver's perception) and inter-individual heterogeneity (differences across drivers). The heterogeneity of the former can be revealed by the variance returned by the perception encoder $[f_\theta(\mathbf{x}_n)]$ in the model pipeline, and the latter can be identified by the mean travel time returned by the encoder. While travelers D and E utilized the same observed route, their simulated alternative routes are disjoint. It is not easy to separate causes of this different perception into intra- and inter-individual heterogeneity using the conventional discrete choice modeling (Bhat and Sardesai, 2006; Krueger et al., 2021). The proposed model captures these differences in perception even when revealed behaviors (=the chosen routes) are identical.

Empirically from the simulations, the generated routes are not arbitrarily dispersed across the network. Despite the Monte Carlo sampling being applied to the entire network, the model does not generate irrational alternatives far from the OD corridor. This suggests that the decision-focused learning process imposes inherent rational constraints, ensuring that the generated stochasticity remains behaviorally plausible. Overall, the model not only reproduces divergent routes across the population but also accounts for varying degrees of route adherence and confidence, among travelers sharing the same spatial and temporal trip backgrounds.

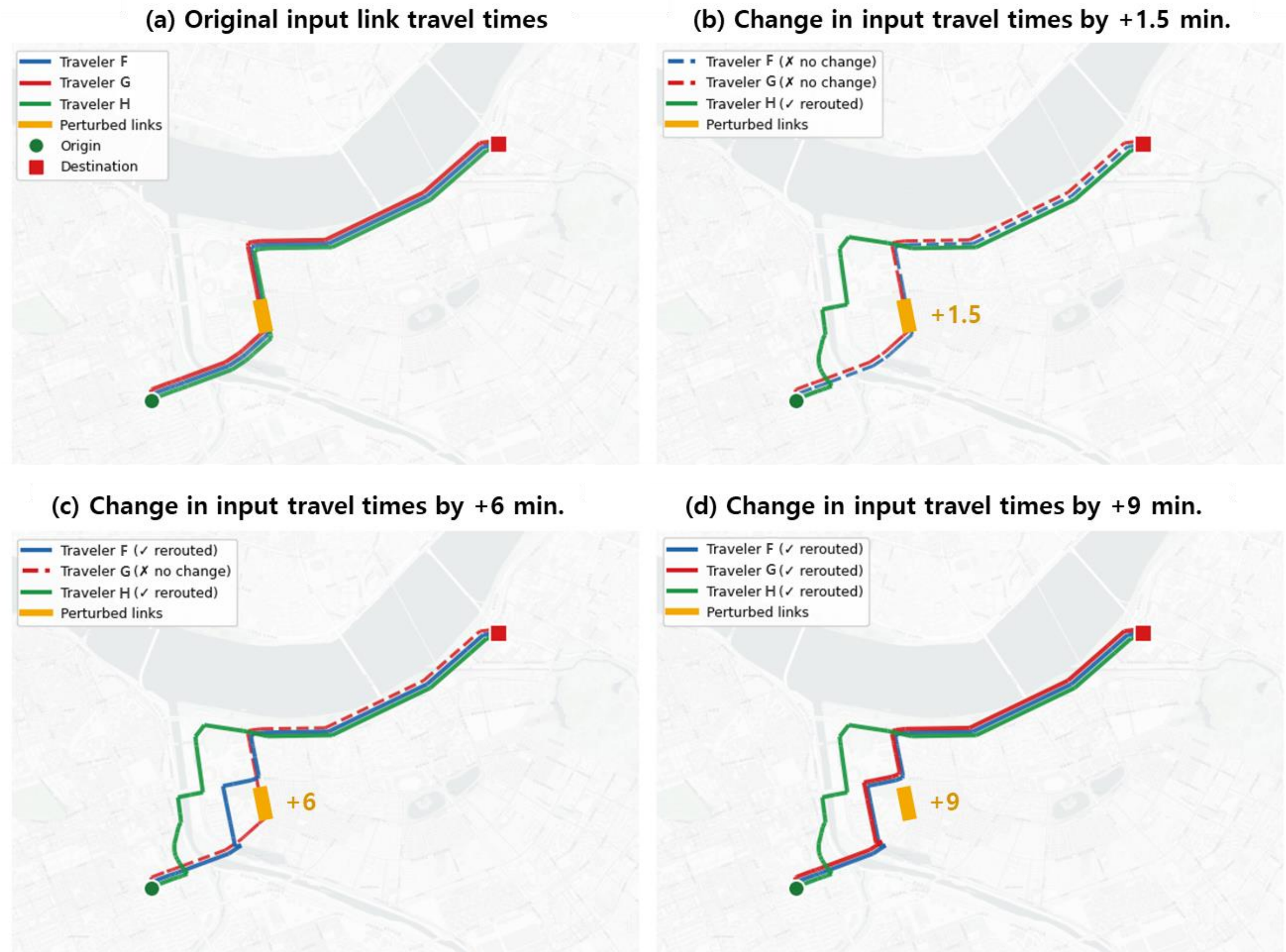

Fig. 13 Counterfactual test for rerouting under perturbed link travel times

To evaluate the model's responsiveness to dynamic changes in network states, we conducted a counterfactual test by systematically increasing the input travel times on a specific road segment. This analysis demonstrates how the proposed model, much like conventional route choice models, adapts its predictions to varying traffic conditions while preserving individual-specific behavioral properties. For this experiment, we selected three travelers (F, G, and H) who share the same OD pair and initially utilize the same observed route, as shown in Fig. 13(a).

The simulation results reveal that travelers possess distinct thresholds for rerouting under identical network disruptions. As illustrated in Fig. 13(b), a modest increase in the input travel time to 1.5 minutes triggers a reroute only for traveler H, while travelers F and G remain on their original paths. When the impedance is further increased to 6 minutes [Fig. 13(c)], traveler F opts for an alternative route, yet traveler G continues to adhere to the original route. Traveler G only deviates from the initial route once the travel time reaches a significant threshold of 9 minutes [Fig. 13(d)]. These findings indicate that even among drivers with identical outward environments, their underlying "resistance" to network changes varies substantially, reflecting diverse levels of risk aversion or route familiarity.

Beyond the decision of whether to reroute, Fig. 13 highlights significant differences in the spatial patterns of reroute. Traveler H exhibits high sensitivity, initiating a detour near the origin, far from the disrupted segment, to proactively avoid the bottleneck. In contrast, travelers F and G demonstrate lower sensitivity, sticking to the original corridor as long as possible and diverting only in the immediate vicinity of the affected links.

The results of this counterfactual test confirm that the proposed framework functions as a robust, general-purpose route choice model capable of responding to varying traffic conditions. Its ability to predict traveler-specific responses, addressing who reroutes, when (at what travel time threshold), and where (the extent of spatial deviation), offers significant practical value. In real-world scenarios such as traffic incidents, road construction, or special events that cause sudden travel time surges, the model can forecast heterogeneous rerouting behaviors. This level of granularity is essential for developing personalized traffic management strategies and for generating traffic information enroute when an unexpected event occurs.

# 6. Conclusions

This study introduces a robust framework to accurately reproduce observed trajectories at the individual level and bypasses path enumeration and overlapping problems. The integration of decision-focused learning facilitates a seamless alignment between the solution of a shortest-path finding problem and observed human behavior. This approach leverages modern machine learning techniques to enable the end-to-end calibration of latent behavioral parameters.

The empirical results demonstrate that the proposed framework consistently outperforms five baseline models in path reproduction. Its superiority is particularly evident in scenarios characterized by high route diversity, where the model excels at identifying and replicating heterogeneous choice patterns across different OD pairs. Furthermore, the model exhibits remarkable robustness against trip scale, whereas the performance of conventional models typically deteriorates as path length and travel time increase, the predictive capability of the proposed framework is reinforced, maintaining high accuracy across long-distance paths.

A significant capability of the proposed model is to retrieve latent perceived travel times. By externalizing these subjective costs, the model provides a microscopic lens through which the underlying motives of route choice can be interpreted. Simulation experiments revealed that travelers under identical traffic conditions arrive at divergent decisions due to their unique, individualized perception fields. Moreover, the analysis of these retrieved outputs confirms that a traveler’s cognitive range is spatially bound, focusing primarily on the immediate trip corridor. This offers insights into the hidden cognitive mechanisms that drive urban mobility.

The primary constraint of the present study lies in the absence of socio-demographic covariates for a more granular retrieval of personalized travel times. The integration of explicit individual-specific data in future research is expected to dramatically enhance the model’s explanatory power and predictive precision. The proposed framework, at its current state, is only for improving navigation services by individualizing the route guidance. The framework needs to be extended to a traffic assignment model that can be used in downstream transportation planning tasks.

## Acknowledgments

This research was supported by the Chung-Ang University research grant in 2026 and by the National Research Foundation of Korea (NRF) Grant funded by the Korean Government (RS-2024-00337956).

# Appendix

## Appendix A. Hyperparameters and experimental settings

Table A1. Hyperparameter settings for training

| Batch size ($B$) | Perturbation scale ($\lambda$) | Weight for KL-divergence ($\xi$) | Perturbation sample size ($s$) | Learning rate ($\eta$) | SoG parameters | | |
|---|---|---|---|---|---|---|---|
| | | | | | $\kappa$ | $\tau$ | $g$ |
| 500 | 0.8 | 0.01 | 1 | 1.25 | 50 | 0.2 | 10 |

In implementation, $\kappa$ was fixed to an empirical constant reflecting the average number of links in the observed routes, for computational simplicity.

Table A2. Constant values and dimensions of covariates

| Testbed network size | | Size of training set ($N$) | Size of test set | Size of sub-test set for diversity |
|---|---|---|---|---|
| Number of nodes ($\|V\|$) | Number of directed links ($\|A\|$) | | | |
| 7,723 | 14,769 | 105,434 | 26,358 | 3,668 |

| **Covariate group** | **Variable definition** |
|---|---|
| Number of individual-specific variables (=driving habit variables) ($d_{person}$) | Summary statistics of driving-habit variables, including the mean, standard deviation, and interquartile range of acceleration, deceleration, and jerk (9) |
| Number of origin-specific location variables ($d_o$) | Building floor area within a 500 m radius of the origin node, classified into 11 building-use categories: residential, business/commercial, neighborhood living, cultural/educational, religious, accommodation/leisure, industrial/production, transportation/automobile, agricultural/livestock/environmental, public, and special-purpose facilities; and the x- and y-coordinates of the origin node (13) |
| Number of destination-specific variables ($d_d$) | Building floor area within a 500 m radius of the destination node, classified into the same 11 building-use categories; and the x- and y-coordinates of the destination node (13) |
| Number of static network states ($d_{static}$) | Link-level attributes, including travel time, link length, number of lanes, road class, and x- and y-coordinates (6) |
| Number of dynamic network states | Mean and standard deviation of link travel time measured at 1hour intervals (2) |
| ($d_{dynamic}$) | |

## Appendix B. Architecture of Perception Sub-Models

| | Traveler attributes encoder<br>($g_{\theta_1}$) | Trip context encoder<br>($g_{\theta_2}$) | Traffic network encoder<br>($g_{\theta_3}$) | Perception encoder<br>($g_{\theta_4}$) |
|---|---|---|---|---|
| Input | $\boldsymbol{x}_n^{person}[B, d_{person}]$ | $\boldsymbol{x}_n^{depT}[B, 1]$,<br>$\boldsymbol{x}_n^{o}[B, d_o]$,<br>$\boldsymbol{x}_n^{d}[B, d_d]$,<br>$\boldsymbol{h}_n^{person}[B, 32]$ | $\boldsymbol{x}_n^{static_net}[B, \lvert A\rvert, d_{static}]$,<br>$\boldsymbol{x}_n^{dynamic_net}[B, \lvert A\rvert, d_{dynamic}]$ | $\boldsymbol{h}_n^{person}[B, 32]$,<br>$\boldsymbol{h}_n^{trip}[B, 256]$,<br>$\boldsymbol{h}_n^{net}[B, \lvert A\rvert, 256]$ |
| Layers | [Input: $\boldsymbol{x}_n^{person}$]<br>Linear (64)<br>LayerNorm&GELU<br>Linear (32)<br>[Output: $\boldsymbol{h}_n^{person}$] | **Temporal encoder**<br>[Input: $\boldsymbol{x}_n^{depT}$]<br>Linear (256)<br>[Output: $Emd_{hour}$]<br><br>**Origin encoder**<br>[Input: $\boldsymbol{x}_n^{o}$]<br>Linear (256)<br>LayerNorm&GELU<br>[Output: $Emd_o$]<br><br>**Destination encoder**<br>[Input: $\boldsymbol{x}_n^{d}$]<br>Linear (256)<br>LayerNorm&GELU<br>[Output: $Emd_d$]<br><br>**OD fusion**<br>[Input: $Emd_o, Emd_d$]<br>Concatenate ($Emd_o, Emd_d$)<br>Linear (256)<br>LayerNorm&GELU<br>[Output: $Emd_{od}$]<br><br>**Personal fusion**<br>[Input: $Emd_{hour}$, $Emd_{od}$, $\boldsymbol{h}_n^{person}$]<br>Concatenate ($Emd_{hour}$, $Emd_{od}$, $\boldsymbol{h}_n^{person}$) $[B, 544]$<br>Linear (256)<br>LayerNorm&GELU<br>Linear (256)<br>LayerNorm&GELU<br>[Output: $\boldsymbol{h}_n^{trip}$] | **Static encoder**<br>[Input: $\boldsymbol{x}_n^{static_net}$]<br>Linear (256)<br>LayerNorm&GELU<br>GCN*3<br>Linear (256)<br>LayerNorm&GELU<br>[Output: $Emd_{static}$]<br><br>**Dynamic encoder**<br>[Input: $\boldsymbol{x}_n^{dynamic_net}$]<br>Linear (256)<br>LayerNorm&GELU<br>GCN*3<br>Linear (256)<br>LayerNorm&GELU<br>[Output: $Emd_{dynamic}$]<br><br>**Fusion**<br>[Input: $Emd_{static}$, $Emd_{dynamic}$]<br>Concatenate ($Emd_{static}$, $Emd_{dynamic}$) $[B, \lvert A\rvert, 512]$<br>Linear (256)<br>LayerNorm&GELU<br>Linear (256)<br>LayerNorm&GELU<br>[Output: $\boldsymbol{h}_n^{net}$] | **Fusion**<br>[Input: $\boldsymbol{h}_n^{person}$, $\boldsymbol{h}_n^{trip}$, $\boldsymbol{h}_n^{net}$]<br>Concatenate ($\boldsymbol{h}_n^{person}$, $\boldsymbol{h}_n^{trip}$, $\boldsymbol{h}_n^{net}$) $[B, \lvert A\rvert, 544]$<br>Linear (256)<br>LayerNorm&GELU<br>Linear (256)<br>LayerNorm&GELU<br>[Output: $Emd_{fusion}$]<br><br>**Mean head**<br>[Input: $Emd_{fusion}$]<br>Linear (256)<br>LayerNorm&GELU<br>Linear (1)<br>[Output: $\mu_n$]<br><br>**Variance head**<br>[Input: $Emd_{fusion}$]<br>Linear (256)<br>LayerNorm&GELU<br>Linear (1)&Softplus<br>[Output: $\Sigma_n$] |
| Output | $\boldsymbol{h}_n^{person}[\mathrm{B}, 32]$ | $\boldsymbol{h}_n^{trip}[B, 256]$ | $\boldsymbol{h}_n^{net}[B, \lvert A\rvert, 256]$ | $\mu_n[B, \lvert A\rvert], \Sigma_n[B, \lvert A\rvert]$ |

## Appendix C. Model specifications of reference models

Table C1. Model specifications of reference models

| Model | Formulation | Input level | Input variables |
|---|---|---|---|
| **PSL** | Discrete choice model with a linear utility specification | Path | Path-level aggregates of link travel time, link length, number of lanes, road class (4) |
| **DNN-PSL** | Discrete choice model with a nonlinear utility specification | Path | Path-level aggregates of link travel time, link length, number of lanes, road class (4) |
| **ReL** | Discrete choice model with a linear utility specification | Link | Link travel time, link length, number of lanes, and road class (4) |
| **AIRL** | Sequential decision model with a nonlinear reward specification | Link | State variables: link travel time, link length, number of lanes, and road class (4) |
| | | | Context variable: departure time and remaining distance to the destination (2) |
| **PURC** | Static optimization | Link | Link travel time, link length, number of lanes, and road class (4) |